\documentclass[authoryear,final,3p,times,twocolumn]{elsarticle}

\usepackage{amssymb}
\usepackage{amsthm}
\usepackage{amsmath}
\usepackage{hyperref}
\usepackage{booktabs}

\usepackage{caption}
\usepackage{algorithm}
\usepackage{algpseudocode}

\graphicspath{{./fig/}}
\usepackage{epsfig} 
\usepackage{graphics} 
\usepackage{svg} 
\usepackage{url}

\journal{Engineering Applications on Artificial Intelligence}
\doi{10.1016/j.engappai.2025.111310}

\begin{document}

\begin{frontmatter}

\title{A real-time anomaly detection method for robots based on a flexible and sparse latent space}

\author[inst1]{Taewook Kang}
\ead{twkang43@hanyang.ac.kr}
\affiliation[inst1]{organization={Department of Computer Science, Hanyang University},
            addressline={Wangsimni-ro 222, Seongdong-gu}, 
            city={Seoul},
            postcode={04763}, 
            country={South Korea}}
            
\author[inst2]{Bum-Jae You}
\ead{ybj@kist.re.kr}
\author[inst2]{Juyoun Park\corref{cor1}\fnref{label1}}
\ead{juyounpark@kist.re.kr}
\author[inst2]{Yisoo Lee\corref{cor1}\fnref{label1}}
\ead{yisoo.lee@kist.re.kr}
\fntext[label2]{Co-corresponding Authors}
\fntext[label3]{This work was supported by the National Research Foundation of Korea (NRF) grant funded by the Korea government(MSIT) (2020M3H8A1114928, No. 2022M3C1A3098746).}

\affiliation[inst2]{organization={Korea Institute of Science and Technology (KIST)},
            addressline={Hwarang-ro 14-gil 5, Seoungbuk-gu}, 
            city={Seoul},
            postcode={02792}, 
            country={South Korea}}

\begin{abstract}
The growing demand for robots to operate effectively in diverse environments necessitates the need for robust real-time anomaly detection techniques during robotic operations.
However, deep learning-based models in robotics face significant challenges due to limited training data and highly noisy signal features.
In this paper, we present Sparse Masked Autoregressive Flow-based Adversarial AutoEncoder model to address these problems.
This approach integrates Masked Autoregressive Flow model into Adversarial AutoEncoders to construct a flexible latent space and utilize Sparse autoencoder to efficiently focus on important features, even in scenarios with limited feature space.
Our experiments demonstrate that the proposed model achieves a 4.96\% to 9.75\% higher area under the receiver operating characteristic curve for pick-and-place robotic operations with randomly placed cans, compared to existing state-of-the-art methods.
Notably, it showed up to 19.67\% better performance in scenarios involving collisions with lightweight objects.
Additionally, unlike the existing state-of-the-art model, our model performs inferences within 1 millisecond, ensuring real-time anomaly detection.
These capabilities make our model highly applicable to machine learning-based robotic safety systems in dynamic environments.
The code is available at \url{https://github.com/twkang43/sparse-maf-aae}.
\end{abstract}




\end{frontmatter}


\section{Introduction}
\label{sec:intro}
In contemporary times, robots are expected to move beyond repetitive tasks in structured factory settings and perform complex tasks in variable environments.
However, the uncertain and variable nature of environments in which robots operate can lead to task failures or malfunctions during operations.
In particular, the growing presence of robots in shared human environments increases the risk of critical accidents from anomalies.
Therefore, promptly detecting and responding to anomalies is essential for the safe and efficient utilization of robots, thereby enhancing the reliability of robotic automation.
To achieve this, the development of anomaly detection methods is essential.
Specifically, for commercialized robots, a real-time anomaly detection method is necessary for practical applicability in real-world scenarios.
In this context, real-time refers to the model's ability to complete inference -- such as signal reconstruction and anomaly scoring -- before the next sliding window is formed from streaming sensor signals, enabling continuous and timely anomaly detection.
These scenarios are further characterized by limited sensors that only measure joint angles or currents through proprioceptive sensors, restricted computational capabilities of embedded boards, and low communication speeds.

There are three approaches for real-time anomaly detection for robots: model-based, knowledge-based, and data-driven \citep{Khalastchi2018}.
Various traditional methods employ model-based approaches, relying on a set of analytical equations and logical formulas to depict normal system behavior \citep{Haddadin2017, Hornung2014, Khalastchi2018}. However, these methods are highly dependent on specific configurations during the threshold-setting process and the accuracy of the model \citep{Haddadin2017}.
Knowledge-based approaches utilize a human expert's behavior to detect anomalies.
They construct topologies and expert systems of the robot frameworks, where each node represents robot components and known fault symptoms. 
Each node stores a value, such as sensor readings or actuator feedback, and a limit that defines the threshold beyond which the node is considered faulty.
Anomalies are detected based on these predefined rates for each node's signals.
These methods interpret anomalies by analyzing the relationships between the nodes where errors are detected \citep{Khalastchi2018}.
However, such approaches have the drawback of being unable to detect anomalies that are not modeled \citep{Hornung2014}.
Both approaches require modeling or empirical settings, which limit anomaly detection in various undefined situations.
Data-driven approaches, on the other hand, rely heavily on data collected during robot operations. They can detect anomalies without specifying models, which is why they are called ``model-free" \citep{Hornung2014, Khalastchi2018}.
Due to this feature, data-driven approaches can be adopted in highly complex systems and detect unknown faults.
There are two types of data-driven approaches: statistical methods and machine learning methods \citep{Khalastchi2018}.
Recently, machine learning methods such as neural networks have mainly been utilized since they can handle complex data patterns effectively, scale well with large datasets, and offer flexibility and adaptability.

Neural network-based time-series anomaly detection approaches have been actively researched in recent times \citep{Pang2021}.
Unsupervised learning, which uses only normal data while training the model, is predominantly used in this field \citep{Pang2021} since the occurrence of anomalies is very rare and significant effort is required to identify and label data.
Models such as Long Short-Term Memory-based Variational AutoEncoder (LSTM-VAE) \citep{Park2018}, MultiVariate Time-series Flow (MVT-Flow) \citep{Brockmann2024}, Anomaly transformer \citep{Xu2022} are proposed to efficiently detect anomalies using unsupervised learning.

The LSTM-VAE uses Variational AutoEncoder (VAE) \citep{Kingma2022} to capture hidden patterns in data and combines them with Long Short-Term Memory (LSTM) networks \citep{Hochreiter1997} for time-series data.
However, when the size of input features is small, VAE struggles to build an efficient latent space since the model typically compresses the input space to construct the latent space.
Additionally, VAE often assumes Gaussian distributions for the posterior and prior, which may not represent the intricate nature of the data, leading to lower anomaly detection rates when data are rare or not well-represented by a normal distribution.
In other words, LSTM-VAE is limited in its application to robots, as the assumption of Gaussian prior and posterior distributions may not accurately represent the robot's actual noisy data distribution \citep{Rezende2015}.

The MVT-Flow utilizes the likelihood of the input data calculated from the Real-valued Non-Volume Preserving (RealNVP) \citep{Dinh2017} architecture, a type of Normalizing Flow (NF) \citep{Rezende2015}, to detect anomalies.
NF transforms simple distributions into complex ones through a series of invertible mappings.
Although NF-based models can accurately mimic the actual data distribution, RealNVP models often face difficulties detecting out-of-distribution data, i.e. anomalies, due to their inductive biases \citep{Kirichenko2020}.
Specifically, they tend to assign higher likelihoods to out-of-distribution data they have not encountered before, making it challenging to distinguish whether the data is anomalous or not.
This means that when MVT-Flow receives an out-of-distribution input, i.e. an anomalous input, it can miscalculate the likelihood of the data and reduce model accuracy.
Therefore, MVT-Flow may struggle to provide robust anomaly detection performance under various conditions, especially when encountering data that is entirely outside the trained distribution.

Anomaly transformer achieves success in the field of anomaly detection by utilizing transformer encoders \citep{Vaswani2017} to capture both temporal and global time-series features of the data.
It can effectively detect anomalies because the transformer can capture global time-series information that recurrent neural networks are unable to retain.
However, due to the algorithm's time and space complexity of $O(L^2N)$, where $L$ represents the length of the time-series sequence and $N$ denotes the number of features, it requires substantial computational time and memory space.
This becomes a drawback when performing real-time anomaly detection with long time steps. 
In robotics, the high computational cost of the method is a critical weakness since it has to be implemented on the embedded computing unit for real-time anomaly detection.
Specifically, using numerous sensor signals increases the computational load, rendering it unsuitable for use on robots, particularly low-cost commercial robots \citep{Yen2019}.
Additionally, the model's large size requires extensive datasets for training, but building these datasets through robot operation is challenging due to the time-intensive data collection process. \citep{Soori2023}.

In this paper, we propose a real-time multivariate time-series anomaly detection model, \textit{Sparse Masked Autoregressive Flow-based Adversarial AutoEncoder} (Sparse MAF-AAE), designed to address the limitations of conventional anomaly detection models.
First, we enhance the adaptability of the model's latent space for datasets, designing it to have a flexible posterior and prior distributions. 
This flexibility is achieved by transforming the posterior and prior distributions within Adversarial AutoEncoder (AAE) \citep{Makhzani2016} using Masked Autoregressive Flow (MAF) \citep{Papamakarios2017}, a type of NF.
The combination of AAE and MAF enables the model to learn a latent space that captures intricate data structures, performing well even with a small amount of training data.
Additionally, this approach enhances robustness against out-of-distribution data by integrating Masked Autoencoder for Distribution Estimation (MADE) \citep{Germain2015} within MAF, utilizing a bottleneck architecture \citep{Kirichenko2020}.
Second, to preserve essential signal characteristics when the number of signal features is small, we incorporate the concept of Sparse autoencoder \citep{Ng2011} by creating a latent space larger than the input while applying constraints that limit parameter updates.
This configuration enables logical compression of the input data and ensures effective feature preservation for constructing an appropriate latent space from sparse input sensor signals.
As a result, it enhances the model's effectiveness in detecting anomalies in robotic operations with limited sensor signals.
Finally, we demonstrated that our methods are feasible for real-time multivariate anomaly detection in robots by constructing models with simple architecture, achieving suitable elapsed times for real-time inferences.
To the best of our knowledge, this is the first work to integrate a flow-based model, AAE, and Sparse autoencoder within a unified framework for signal reconstruction-based anomaly detection.

Comparative experiments are conducted to evaluate the real-time multivariate anomaly detection performance based on a modified version of the voraus-AD dataset \citep{Brockmann2024}.
The dataset offers time-series signals of the manipulator when a pick-and-place operation is performed with 12 anomaly types caused by heterogeneous factors.
To implement the real-time detection, the sliding window is applied for the implemented anomaly detection methods.
Results showed that our model achieved the highest \textit{area under the ROC curve} (AUROC) score among the comparison models, while also maintaining an appropriate computational time for real-time multivariate time-series anomaly detection in robotic applications. 

\section{Background Theory}\label{sec:background}

\subsection{Masked Autoregressive Flow}\label{sec:backgroud:maf}
MAF \citep{Papamakarios2017} is an NF that leverages the autoregressive properties inherent in the data.
NF is a network that transforms a probability density through a sequence of invertible mappings $f_1, f_2, \cdots, f_K$ \citep{Rezende2015}.
The transformed sample space $\mathbf{z}_K$ with distribution $q_K(\mathbf{z}_K)$ is obtained by applying a series of functions $f_k$ to the initial sample space $\mathbf{z}_0$.
\begin{equation}\label{eq:zk}
    \mathbf{z}_K = f_K \circ \cdots \circ f_2 \circ f_1(\mathbf{z}_0).
\end{equation}
The log-likelihood of the transformed sample space $\mathbf{z}_K$ is represented by the following expression:
\begin{equation}\label{eq:lnzk}
    \ln q_K(\mathbf{z}_K) = \ln q_0(\mathbf{z}_0)-\sum_{k=1}^K \ln \left| \det{\frac{\partial f_k(\mathbf{z}_{k-1})}{\partial \mathbf{z}_{k-1}}} \right|.
\end{equation}
Equation (\ref{eq:lnzk}) is used to optimize NF models based on maximum likelihood estimation.
Each mapping $f_k$ should not only be invertible but also have an easily computable Jacobian determinant to reduce the computational load.

MAF transforms the autoregressive input data $\mathbf{x}$ into random variables $\mathbf{u}$ using a sequence of transformations $f$, and then computes the probability density $p(\mathbf{x})$ from the transformed density $p(\mathbf{u})$ through the inverse transformations $f^{-1}$.
The relationship between $x_i$ and $u_i$, which are $i^{\text{th}}$ components of $\mathbf{x}$ and $\mathbf{u}$ respectively, can be expressed as follows:
\begin{equation}\label{eq:mafxi}
    x_i = u_i \exp(\alpha_i) + \mu_i,
\end{equation}
where $\mu_i=f_{\mu_i}(\mathbf{x}_{1:i-1})$, $\alpha_i=f_{\alpha_i}(\mathbf{x}_{1:i-1})$, and $u_i \sim \mathcal{N}(0,1)$.
Here, $f_{\mu_{i}}$ and $f_{\alpha_i}$ are unconstrained scalar functions that compute mean $\mu_i$ and log standard deviation $\alpha_i$ of the $i^{\text{th}}$ conditional given all previous variables.
In \citep{Papamakarios2017}, the authors employed MADEs \citep{Germain2015} to derive $\mu_i$ and $\alpha_i$.
This approach offers the benefits of enforcing the autoregressive property and being computationally efficient for inverse operations on equation (\ref{eq:mafxi}).
Using equation (\ref{eq:mafxi}), the entire process can be expressed as follows:
\begin{equation}\label{eq:mafx}
    \mathbf{x}=f(\mathbf{u})
    \quad \text{where} \quad \mathbf{u} \sim \mathcal{N}(\mathbf{0}, \mathbf{I}).
\end{equation}
The above formula \eqref{eq:mafxi} and \eqref{eq:mafx} can be modified in the autoregressive density transformation process as follows:
\begin{equation}\label{eq:mafzi}
    z_{k_i}=z_{k_{i-1}}\exp(\alpha_i)+\mu_i,
\end{equation}
\begin{equation}\label{eq:mafzk}
    \mathbf{z}_k=f(\mathbf{z}_{k-1}),
\end{equation}
where $\alpha_i = f_{\alpha_i}(\mathbf{z}_{k_{1:i-1}})$, $\mu_i = f_{\mu_i}(\mathbf{z}_{k_{1:i-1}})$, and $\textbf{z}_k \sim q_k(\mathbf{z}_k)$.

By equations (\ref{eq:mafzi}) and (\ref{eq:mafzk}), the base sample space $\mathbf{z}_0$ can be transformed into complex sample space $\mathbf{z}_K$ by equation (\ref{eq:zk}) using a set of $\{\mathbf{\alpha}_k, \mathbf{\mu}_k\}$:
\begin{equation}\label{eq:fk}
    f_{k}(\mathbf{z}_{k-1}) = \mathbf{z}_{k-1} \odot \exp(\boldsymbol{\alpha}_k)+\boldsymbol{\mu}_k,
\end{equation}
where $\boldsymbol{\alpha}_k = (\alpha_1, \alpha_2, \cdots, \alpha_K)$, $\boldsymbol{\mu}_k = (\mu_1, \mu_2, \cdots, \mu_K)$, and $\odot$ denotes element-wise multiplication.

\subsection{Adversarial AutoEncoder}\label{sec:background:aae}
The AAE is a sophisticated generative model that integrates concepts from autoencoders and adversarial training \citep{Makhzani2016}.
This model performs variational inference by matching the aggregated posterior of the latent code vector with an arbitrary prior distribution.
Although it shares conceptual similarities with the VAE \citep{Kingma2022}, the AAE differs in its approach to measuring the discrepancy between the posterior $q(\mathbf{z})$ and the prior $p(\mathbf{z})$.
In particular, VAEs typically compute the Kullback-Leibler (KL) divergence between $q(\mathbf{z})$ and $p(\mathbf{z})$, incorporating this as a regularization term.
To calculate the divergence analytically, this approach typically requires that the prior distribution be a standard normal distribution and the posterior distribution be explicitly defined as Gaussian.
Conversely, the AAE leverages Generative Adversarial Networks (GANs) \citep{Goodfellow2014} to approximate the target distribution of $p(\mathbf{z})$.
Through adversarial training, the AAE effectively minimizes the discrepancy between $q(\mathbf{z})$ and the target distribution.
Thus, the explicit equations of the two distributions are not required to construct a closed-form solution, which permits the distributions to assume a more complex form.

The AAE framework comprises two distinct networks: the generator $G$ and the discriminator $D$.
Within the context of adversarial learning, $G$ operates as the encoder in the autoencoder.
$G$ aims to generate an aggregated posterior distribution $q(\mathbf{z})$ that can successfully deceive $D$ into believing that the latent code from $q(\mathbf{z})$ is sampled from the true prior distribution $p(\mathbf{z})$.
$D$ accurately distinguishes between the true samples from $p(\mathbf{z})$ and the generated samples from $q(\mathbf{z})$.
In the reconstruction phase, the AAE updates both encoder and decoder to minimize the reconstruction error of the input data.
During the regularization phase, the model is trained by optimizing the performance of both networks, $G$ and $D$, to ensure that $q(\mathbf{z})$ closely approximates $p(\mathbf{z})$.
After training, the model can generate outputs by decoding from the latent code sampled from the learned distribution $q(\mathbf{z})$, which mimics $p(\mathbf{z})$.

\subsection{Sparse autoencoder}\label{sec:background:sae}
Sparse autoencoder \citep{Ng2011} is a type of autoencoders that applies mathematical constraints on the network to construct sparse but meaningful hidden representations.
Whereas traditional autoencoders typically limit the number of hidden units to extract essential features, Sparse autoencoder imposes a sparsity constraint that permits more hidden units than input units, ensuring that most units remain inactive.
This approach enables the model to logically extract important structures in the data, as only a few important units are activated, even with a larger hidden layer.

Sparse autoencoders can utilize a variety of sparsity constraints, such as k-sparse autoencoders \citep{Makhzani2014}, $L_1$ regularization, and KL divergence \citep{Ng2011}.
We employ $L_1$ regularization to implement sparsity constraints for our model:
\begin{equation}\label{eq:l1}
    L_1\text{ regularization} = \mathcal{L}(x,\hat{x}) + \lambda\sum_{i}\|w_i\|_1,
\end{equation}
where $\mathcal{L}(x,\hat{x})$ is the reconstruction error, $\lambda$ is a hyperparameter, and $w$ represents the model parameters.
$L_1$ regularization tends to drive many parameters to zero, resulting in a sparse parameter matrix \citep{Ng2004}.
Therefore, this regularization imposes constraints on parameter updates, activating only a few hidden units to construct a compressed representation of the input data.

\section{Methodology}
In this section, we propose Sparse MAF-AAE, a real-time anomaly detection method that reconstructs data by effectively modeling both the posterior and prior distributions, thereby accurately reflecting the characteristics of the input data.
First, an overview of the framework is provided in Section \ref{sec:method:framework_overview}.
Next, in Section \ref{sec:method:sparse_mafaae}, the model architecture is explained in detail.
Finally, we elaborate on the method for training the model and performing real-time anomaly detection in Sections \ref{sec:method:offline_training} and \ref{sec:method:real_time_detection}.
It is recommended to view all figures in this section in color print.

\subsection{Framework Overview}\label{sec:method:framework_overview}
Our model is designed for real-time multivariate time-series anomaly detection during robotic tasks, such as pick-and-place operations.
The model learns a latent space from input data through encoder layers, and then reconstructs the data through a decoder.
The model is trained using unsupervised learning (i.e., trained exclusively on normal data), enabling it to accurately reconstruct only normal data.
Consequently, when anomalous data is presented to the model, it fails to reconstruct it appropriately, since it has not been trained on such data.
Therefore, the difference between the abnormal input data and the reconstructed result may be significant, allowing the detection of errors based on this discrepancy.

The developed framework consists of two stages: Offline Training and Real-time Detection.
During the preprocessing stage of training, the preprocessor generates the input data $W_{\mathbf{x}[t:t+T_W]} \in \mathbb{R}^{T_W \times N}$, such as torque data from the robot, from the complete multivariate time-series task sequence $\mathbf{x} \in \mathbb{R}^{T \times N}$.
Here, $T_W$ denotes the length of each window, $T$ represents the total length of time steps from the beginning to the end of a specific task, and $N$ indicates the number of signals.
In the training procedure, the model is trained to accurately reconstruct the time-series signals within each window.
Once the training for one window is complete, the window shifts to the next data sequence with a stride of $T_S$.
Detailed information on the model architecture and the offline training methodology can be found in Section \ref{sec:method:sparse_mafaae} and \ref{sec:method:offline_training}.

After the model is fully trained, it is used for real-time anomaly detection. 
Before detection, the mean $\mu_{\text{normal}}$ and standard deviation $\sigma_{\text{normal}}$ are calculated based on the reconstruction error of the normal data. 
During real-time anomaly detection, when a new data frame $\mathbf{w}$ is received, the model performs reconstruction and calculates the reconstruction error $L_1\left(\mathbf{w},\mathbf{w}^\prime\right)$.
The reconstruction error is then converted to an anomaly score $AS(\mathbf{w})$ based on the $\mu_{\text{normal}}$ and $\sigma_{\text{normal}}$. Based on the anomaly scores and an arbitrary threshold $\theta$, the system determines whether the current robot operation is anomalous or not.
Section \ref{sec:method:real_time_detection} provides a comprehensive description of the real-time anomaly detection process.

\begin{figure}[t]
    \centering
    \includegraphics[width=\linewidth]{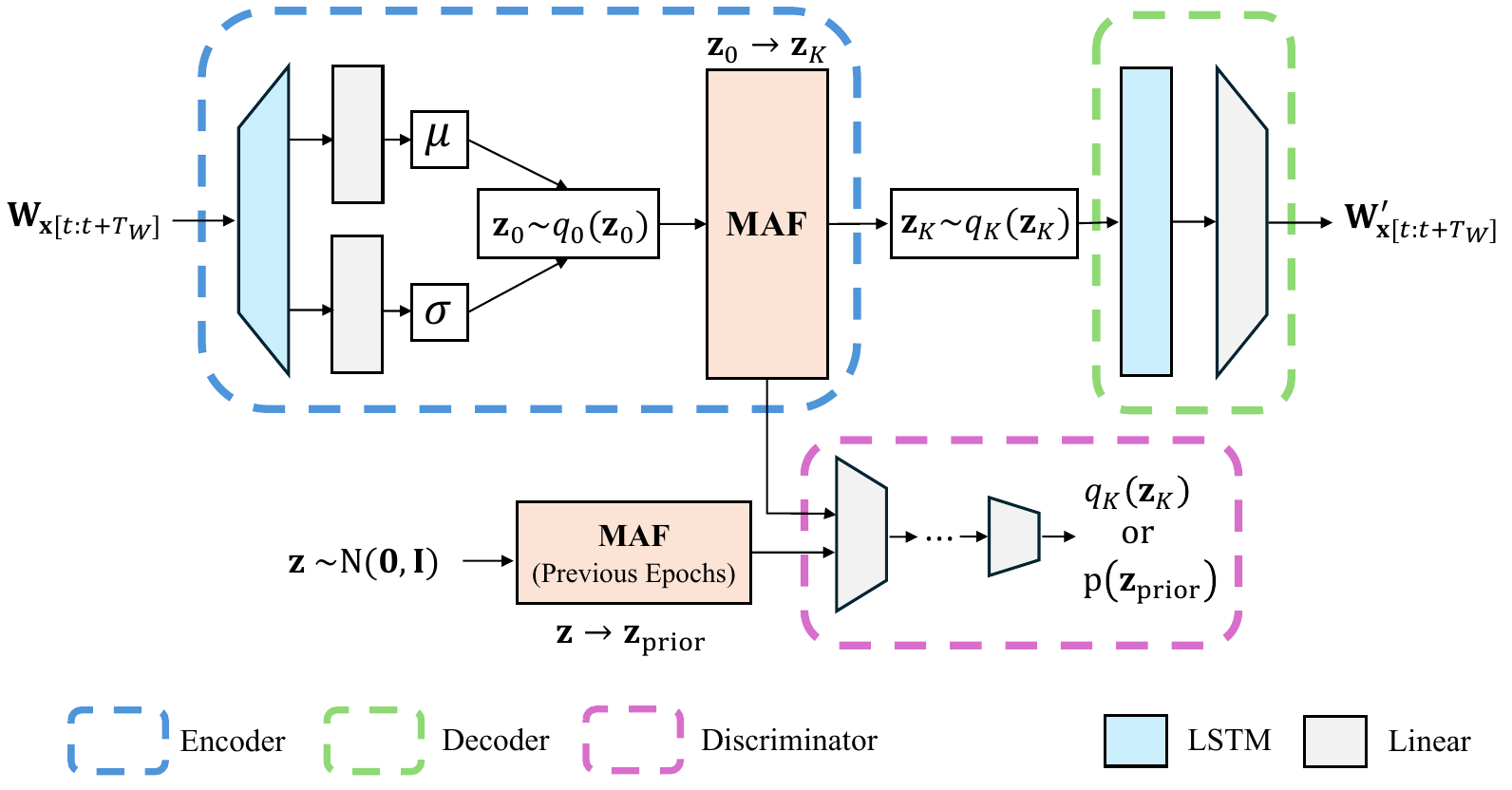}
    \caption{Overall architecture of Sparse MAF-AAE}
    \label{fig:overall_architecture}
\end{figure}

\subsection{Sparse MAF-AAE}\label{sec:method:sparse_mafaae}
Our model, Sparse MAF-AAE, is constructed based on the AAE \citep{Makhzani2016}.
We improve the model by incorporating MAF \citep{Papamakarios2017} layers to enhance the complexity of the latent space generated by the AAE encoder, similar to the approach by \citep{Su2019}, which applied planar NF \citep{Rezende2015} to the VAE encoder \citep{Kingma2022}.
Additionally, we expand the size of both the hidden space and the latent space while imposing sparsity constraints \citep{Ng2011} through $L_1$ regularization \citep{Ng2004} to capture an optimized representation of the data even with a limited number of features.
Figure \ref{fig:overall_architecture} presents the overall architecture of Sparse MAF-AAE.

Unlike VAEs, which constrain posterior and prior distributions to follow a normal distribution to match the distributions, the AAE uses GANs \citep{Goodfellow2014} to align the posterior distribution to an arbitrary prior distribution.
This GAN-based alignment allows the AAE to create flexible posterior and prior distributions without imposing specific mathematical models on the distributions, such as a Gaussian, thus making it more suitable for capturing complex data features.
Leveraging this flexibility, we incorporate an MAF layer - a neural network that transforms a simple initial distribution into a final complex distribution - within the AAE architecture to create adaptable posterior distributions $q_K(\mathbf{z}_K)$ and prior distributions $p(\mathbf{z}_\text{prior})$.
This enhancement allows the model to construct a latent space $\mathbf{z}_K$, sampled from $q_K(\mathbf{z}_K)$, that accurately captures and represents the features of the input data.

In detail, the encoder first constructs a base latent space $\mathbf{z}_0$, which is sampled from a Gaussian distribution $q_0(\mathbf{z}_0)$.
Specifically, we adapt LSTMs \citep{Hochreiter1997} to construct a hidden space, as the input data is time-series data.
The model calculates the mean vector $\boldsymbol{\mu}$ and the standard deviation vector $\boldsymbol{\sigma}$ from hidden space using linear networks to construct a base distribution $q_0(\mathbf{z}_0)$.
Subsequently, it samples the base latent space $\mathbf{z}_0$ using the reparameterization trick \citep{Kingma2022} as follows:
\begin{equation}
    \mathbf{z}_0 = \boldsymbol{\mu} + \boldsymbol{\sigma} \odot \boldsymbol{\epsilon},
    \quad \boldsymbol{\epsilon} \sim \mathcal{N}(\mathbf{0}, \mathbf{I}).
\end{equation}

\begin{figure*}[t]
    \centering
    \includegraphics[width=0.8\linewidth]{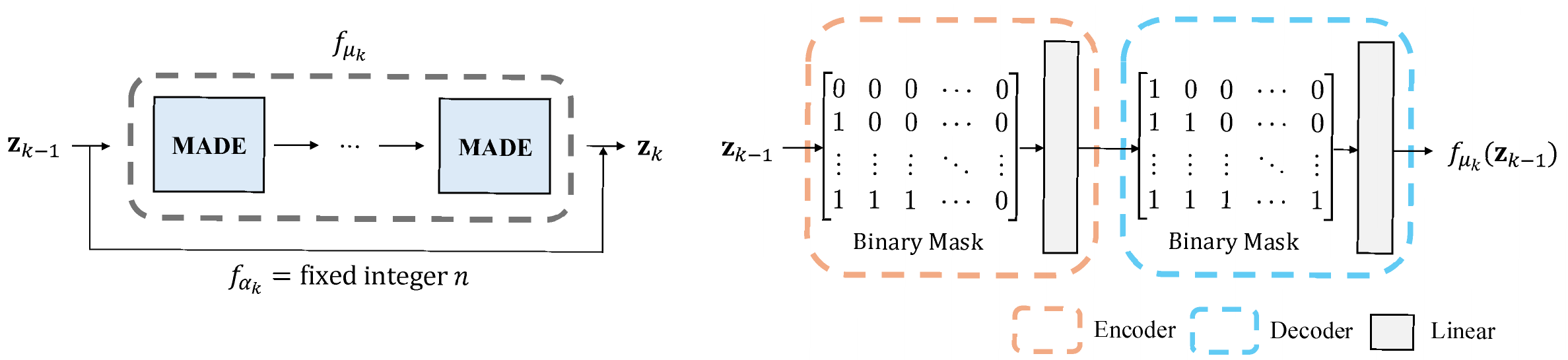}
    \caption{Architecture of the MAF layer (left) and the MADE layer (right)}
    \label{fig:maf_made_architecture}
\end{figure*}

Using MAF, the model then transforms $\mathbf{z}_0$ into $\mathbf{z}_K$, which represents the final latent space containing the salient features of the input data by equation (\ref{eq:zk}) and (\ref{eq:fk}).
To utilize the MAF layer, we enforce the latent space to exhibit autoregressive features by stacking MADEs \citep{Germain2015} on $f_{\mu_k}$.
Figure \ref{fig:maf_made_architecture} illustrates the architecture of the MAF layer and the MADE layer.
MADE is an autoencoder that has masks on its encoder and decoder parameters to ensure autoregressive properties in the data.
These masks enforce autoregressive characteristics by eliminating computational paths, i.e, setting connections to 0, between the output $\hat{z}_{k_d}$ and any of the subsequent inputs $z_{k_d}, z_{k_{d+1}}, \cdots, z_{k_D}$.
To achieve this, the matrix product of the encoder and decoder masks, $\mathbf{M}_{\text{encoder}}\mathbf{M}_{\text{decoder}}$, in the MADEs must be a strictly lower triangular matrix.
We employ a strict lower triangular matrix $\mathbf{M}_{\text{encoder}} \in \{0,1\}$ and a lower triangular matrix $\mathbf{M}_{\text{decoder}} \in \{0,1\}$ as a binary mask for an encoder and a decoder, respectively.
To ensure stable learning, we adopt fixed non-negative integer values for $f_{\alpha_k}$, inspired by residual connections \citep{He2016}.
With this approach, the model can successfully learn a final latent space $\mathbf{z}_K$ that captures the complex distribution of the input data.

In the field of robotics, robust anomaly detection is essential even with limited signals, as adding more signals demands too much computing power for embedded systems \citep{Yen2019}.
Therefore, traditional autoencoders are unsuitable for these cases, as constraining the latent space to be smaller than the compact input space might prevent the model from constructing an appropriate latent representation.
We address this problem by adopting the concept of Sparse autoencoder \citep{Ng2011}.
We doubled the size of the hidden space and the latent space while applying $L_1$ regularization to the model's loss function as a sparsity constraint.
By this approach, the model implements the crucial concept of autoencoders, which is to learn efficient representations of input data by activating important cells.

In Sparse MAF-AAE decoder, the model initially maps the latent space $\mathbf{z}_K$ to a hidden space.
Subsequently, the decoder reconstructs data $W_{\mathbf{x}[t:t+T_W]}^\prime$ from a hidden space through a series of linear layers.
When the model reconstructs data from the latent space generated by normal data, it produces outputs that closely resemble the input data.
However, when encountering anomalous data, the output will likely differ significantly from the input since the model has not encountered it before.
This distinction underscores a key characteristic of our model: its ability to differentiate between normal and abnormal data.
This feature is utilized to detect anomalies in the real-time anomaly detection process, which is comprehensively explained in Section \ref{sec:method:real_time_detection}.
Overall, the integrated system of the encoder and decoder in Sparse MAF-AAE is collectively referred to as the generator $G$.

The discriminator $D$ of Sparse MAF-AAE aligns the posterior $q_K(\mathbf{z})$ to the prior $p(\mathbf{z_\text{prior}})$ by assessing whether the sample space $\mathbf{z}$ came from  $q_K(\mathbf{z}_K)$ or $p(\mathbf{z}_\text{prior})$.
We constructed $D$ using a simple multi-layer perceptron.
With the result from $D$, $G$ is trained to generate $q_K(\mathbf{z})$ that cannot be distinguished from $p(\mathbf{z_\text{prior}})$.
We chose to use $q_K(\mathbf{z})$ from previous learning (e.g., previous epochs, previous batches) as $p(\mathbf{z}_\text{prior})$ for current learning, similar to the concept of updating priors in Bayesian inference \citep{Bishop2006}.
During training, the model constructs an appropriate prior $p(\mathbf{z}_\text{prior})$ and ensures that $q_K(\mathbf{z}_K)$ follows $p(\mathbf{z}_\text{prior})$ through dynamic adjustment of the regularization term.

\subsection{Offline Training}\label{sec:method:offline_training}

\newcommand{\factorial}{\ensuremath{\mbox{\sc Factorial}}}
\begin{algorithm*}[t]
\footnotesize
\caption{Training procedure of Sparse MAF-AAE}\label{pseudo:min_L}
\begin{algorithmic}[1]
\State $\theta_G$ : Parameters of $G$, $\theta_D$ : Parameters of $D$\\

\For{each batch $\mathbf{x}$ in train\_loader}
    \State Initialize lists for $G$ and $D$ updates: $\mathcal{U}_G, \mathcal{U}_D$

    \For{each sample $\mathbf{x}_i$ in batch $\mathbf{x}$}
        \State $\{W_{\mathbf{x}_i[t,t+T_W]}\}_{t=0,\cdots} \leftarrow \texttt{sliding\_window}(\mathbf{x}_i)$

        \For{each window $W_{\mathbf{x}_i[t,t+T_W]}$ in $W_{\mathbf{x}_i[t,t+T_W]}\}_{t=0,\cdots}$}
            \State $W_{\mathbf{x}_i[t,t+T_W]}^\prime, \mu, \log(\sigma^2) = G(W_{\mathbf{x}_i[t,t+T_W]})$\\

            \State $\mathcal{L}_G \leftarrow \texttt{get\_generator\_loss}(W_{\mathbf{x}_i[t,t+T_W]}^\prime, W_{\mathbf{x}_i[t,t+T_W]}, \mu, \log(\sigma^2))$
            \State $\mathcal{U}_G \leftarrow \mathcal{U}_G \cup \{\mathcal{L}_G\}$ : Append  $\mathcal{L}_G$ to $\mathcal{U}_G$\\

            \State $\mathcal{L}_D \leftarrow \texttt{get\_discriminator\_loss}(\mu, \log(\sigma^2))$
            \State $\mathcal{U}_D \leftarrow \mathcal{U}_D \cup \{\mathcal{L}_D\}$ : Append  $\mathcal{L}_D$ to $\mathcal{U}_D$
        \EndFor
    \EndFor

    \State \quad $\theta_G \leftarrow \theta_G - \texttt{AdamW}(\nabla_{\theta_G} \frac{1}{|\mathcal{U}_G|} \sum \mathcal{U}_G)$
        
    \State \quad $\theta_D \leftarrow \theta_D - \texttt{AdamW}(\nabla_{\theta_D} \frac{1}{|\mathcal{U}_D|} \sum \mathcal{U}_D)$
\EndFor

\end{algorithmic}
\end{algorithm*}

As described in Section \ref{sec:method:sparse_mafaae}, Sparse MAF-AAE is composed of two networks: the generator $G$ and the discriminator $D$.
Each networks have its own loss function, $\mathcal{L}_G$ and $\mathcal{L}_D$, respectively.
The model is optimized through iterative updates of the $G$ and $D$ parameters over a specified number of epochs, as delineated in Algorithm \ref{pseudo:min_L}.
The training procedure consists of two steps: First, $G$ is trained by minimizing $\mathcal{L}_G$ to accurately reconstruct the data, achieve sparse activation of the model parameters, and ensure that $q_K(\mathbf{z}_K)$ closely approximates $p(\mathbf{z}_\text{prior})$.
Second, $D$ is trained by minimizing $\mathcal{L}_D$ to effectively discriminate whether the sampled space $\mathbf{z}$ is derived from the posterior $q_K(\mathbf{z}_K)$ or the prior $p(\mathbf{z}_\text{prior})$.

$G$ is trained using the following loss function $\mathcal{L}_G$:
\begin{equation}\label{eq:overall_loss}
    \mathcal{L}_G=
    \mathcal{L}_{\text{MSE}} + 
    \mathcal{L}_{L_1} +
    \mathcal{L}_\text{BCE}.
\end{equation}
The term $\mathcal{L}_{\text{MSE}}$ represents the mean squared error between the input window $W_{\mathbf{x}[t:t+T_W]}$ and the reconstructed window $W_{\mathbf{x}[t:t+T_W]}^\prime$. This error is minimized to enhance the model's reconstruction ability.
\begin{equation}\label{eq:mse}
    \mathcal{L}_{\text{MSE}} = \sum \left(W_{\mathbf{x}[t:t+T_W]} - W_{\mathbf{x}[t:t+T_W]}^\prime\right)^2.
\end{equation}
The term $\mathcal{L}_{L_1}$ represents the sparsity constraint.
As discussed in Section \ref{sec:background:sae} and \ref{sec:method:sparse_mafaae}, $L_1$ regularization restricts model updates, activating only a few cells \citep{Ng2004}.
The $L_1$ regularization term for Sparse MAF-AAE is expressed as follows:
\begin{equation}
    \mathcal{L}_{L_1}=\lambda\sum \|w_{\text{encoder}}\|_1,
\end{equation}
where $w_{\text{encoder}}$ denotes the parameters of the encoder.
By imposing this constraint on the encoder, the model constructs a latent space by encoding the input data using only a few significant cells.
The influence of this term can be adjusted by the hyperparameter $\lambda$.
In AAE models, the model is optimized not only to accurately reconstruct the data but also to follow the true prior distribution of the data’s latent space \citep{Makhzani2016}.
To achieve this, the divergence between the posterior and prior distributions is minimized through the concurrent optimization of both generator and discriminator loss.
Sparse MAF-AAE achieves this by concurrently minimizing $\mathcal{L}_\text{BCE}$ for $G$ and $\mathcal{L}_D$ for $D$.
$G$ is trained by minimizing the following binary-cross entropy loss:
\begin{equation}\label{eq:bce_in_loss}
    \mathcal{L}_\text{BCE} =
    \beta \times \text{BCE}(D(\mathbf{z}_K), \mathbf{1}).
\end{equation}
This term denotes the regularization error in generative autoencoders.
Consequently, the coefficient $\beta$ can precisely control the trade-off between reconstruction accuracy and the ability to follow the desired distribution through hyperparameter tuning, similar to the technique proposed by $\beta-\text{VAE}$ \citep{Higgins2017}.
We incorporated the concept of KL annealing \citep{Liang2018} to adjust the hyperparameter $\beta$, gradually increasing it during the training.
Initially, when $\beta$ is small, the model prioritizes accurate reconstruction; as $\beta$ increases, the model focuses on aligning $q_K(\mathbf{z}_K)$ with $p(\mathbf{z}_\text{prior})$.
The MAF layer is implicitly trained using $\mathcal{L}_\text{MSE}$ and $\mathcal{L}_\text{BCE}$, as these loss functions contribute to constructing transformation functions $f$ that yield an appropriate posterior distribution.

$D$ is optimized by minimizing the following average of binary-cross entropy losses:
\begin{equation}\label{eq:discriminator_loss}
    \mathcal{L}_D = \frac{1}{2}\left[
    \text{BCE}(D(\mathbf{z}_\text{prior}), \mathbf{1}) +
    \text{BCE}(D(\mathbf{z}_K), \mathbf{0})
    \right].
\end{equation}
By training $D$ with $\mathcal{L}_D$, the network effectively discriminates whether an arbitrary latent space $\mathbf{z}$ was sampled from $q_K(\mathbf{z}_K)$ or $p(\mathbf{z}_\text{prior})$.
When implementing this loss in code, $\mathbf{z}_K$ must be detached from the computation graph while minimizing $\mathcal{L}_D$.
This is necessary because, if $\mathbf{z}_K$ is not detached, it would inadvertently affect $G$ during the training of $D$.
Thus, the detachment ought to be applied to ensure stable training.

\subsection{Real-time Detection}\label{sec:method:real_time_detection}
Once Sparse MAF-AAE is fully trained offline, it can be used for real-time anomaly detection.
Similar to the method proposed in \citep{Chen2023}, our model uses historical information to determine whether the input data contains errors.
Before deployment, the system calculates the mean $\mu_\text{normal}$ and standard deviation $\sigma_\text{normal}$ of the $L_1$ norms of a set of normal data for an output normalization.
The formula for calculating the $L_1$ norm for the window-sized input signals $\mathbf{w}$ is as follows:
\begin{equation}\label{eq:l1_norm}
    L_1(\mathbf{w},\mathbf{w}^\prime) = 
    \left\|\mathbf{w} - \mathbf{w}^\prime\right\|_1,
\end{equation}
where the $\mathbf{w}^\prime$ is the reconstruction of $\mathbf{w}$.

Normal data are likely to produce detailed and accurate reconstructed signals, whereas anomalous data will result in reconstructions that do not closely resemble the input.
Therefore, as the data becomes more anomalous, the $L_1$ norm correspondingly increases.
Based on this concept, the system calculates the anomaly score $\text{AS}(\mathbf{w})$ of the input data $\mathbf{w}$ during the real-time anomaly detection phase using $\mu_\text{normal}$, $\sigma_\text{normal}$, and a $L_1$ norm of $\mathbf{w}$. The formula used is as follows:
\begin{equation}\label{eq:anomaly_score}
    \text{AS}(\mathbf{w})=\frac{L_1(\mathbf{w})-\mu_\text{normal}}{\sigma_\text{normal}}.
\end{equation}
\begin{equation}\label{eq:determine_anomaly}
    \text{IsAnomaly}(\mathbf{w})=
    \begin{cases}
        \text{\ true} \quad &\text{if} \quad \text{AS}(\mathbf{w}) > \theta,\\
        \text{\ false} \quad &\text{otherwise}.
    \end{cases}
\end{equation}
The anomaly score (\ref{eq:anomaly_score}) is used to assess if the data contains anomalies based on the threshold $\theta$.
If $\text{AS}(\mathbf{w})$ is bigger than $\theta$, $\mathbf{w}$ is classified as anomalous data.
The threshold $\theta$ can be configured to meet user needs: it can be set manually through user experience or adaptively determined based on anomaly score statistics, or adjusted using other creative criteria that can effectively discriminate between anomalies and normal data.
This flexibility allows users to adjust the threshold according to their specific application requirements.

When legitimate adjustments are made to the robotic platform or its operating environment, retraining the model is necessary to maintain optimal performance.
However, since our model uses unsupervised learning, the retraining process is straightforward: users only need to provide normal operational data to the model and set an appropriate threshold.
This streamlined data requirement enables easy adaptation and redeployment of the model for effective anomaly detection.

\section{Experiments}
In this section, we first provide a detailed description of the dataset used to compare models in Section \ref{sec:experiments:voraus-ad}.
Then, we present our evaluation methodology, the metrics employed, and the implementation details of the models in Section \ref{sec:experiments:eval_method} and \ref{sec:experiments:implem_details}.
The experimental results are shown in Section \ref{sec:experiments:results}.
Finally, in Section \ref{sec:experiments:ablation}, we provide a detailed analysis of the validity of our experiments and the impact of our improvements on the model.

\begin{figure*}[t]
    \centering
    \includegraphics[width=\linewidth]{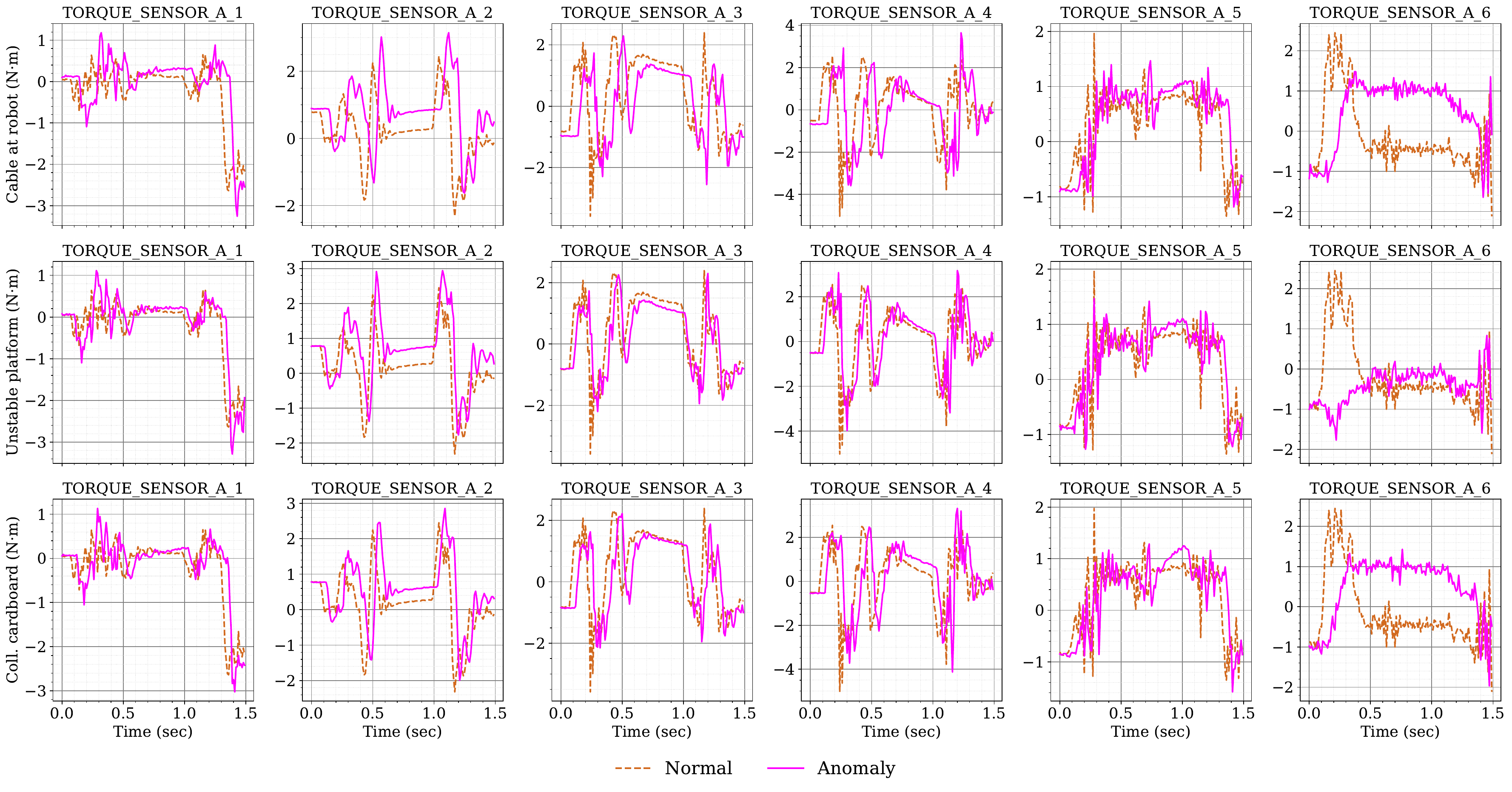}
    \caption{
    Comparison of normal (dashed brown) and anomalous (magenta) torque signals for three anomalies from the 50 Hz dataset over a 1.5-second window. Each row represents a distinct anomaly and each column corresponds to a joint’s torque sensor. The figure shows that anomalies are clearly distinct from normal operation, with each anomaly exhibiting a unique torque profile.}
    \label{fig:anomaly_normal_torques}
\end{figure*}

\subsection{Dataset}\label{sec:experiments:voraus-ad}
In our experiments, we used a customized version of the voraus-AD dataset \citep{Brockmann2024}, tailored to our research needs.
The voraus-AD dataset is designed for anomaly detection in the pick-and-place operations of the collaborative robotic arm \textit{Yu-Cobot}.
The robot performs the operation by using a vacuum gripper to pick up a randomly placed can from a conveyor belt, and subsequently moving it to a fixed target position.
Each record contains a total of 130 signals, such as torques, joint positions, and torque-forming currents.

The dataset consists of 2,122 records, with 948 samples for training and 1,174 for testing.
Within the test dataset, 755 samples are anomalies and 419 samples are normal.
It includes 12 types of anomalies, which can be categorized into three causes: \textit{Processor Errors}, \textit{Gripping Errors}, and \textit{Robot Axis Wear}.
\begin{enumerate}
    \item \textit{Process Errors}: 
    These errors are anomalies induced by the environmental conditions surrounding the robot.
    Collisions of the robot with lightweight objects, and variations in the weight of the can to be gripped are examples.
    \item \textit{Gripping Errors}:
    Defects or contamination of the gripper can cause the robot to malfunction.
    Misgripping the can and losing the can during the operation are classified as \textit{Gripping Errors}.
    \item \textit{Robot Axis Wear}:
    Increased joint friction and the shift of the current curve resulting from motor miscommutation are attributed to wear on the robot axis.
\end{enumerate}
All anomaly types, their corresponding causes, and the number of occurrences for each anomaly are presented in Table \ref{tab:anomalies}.
In our experiment, we used a test dataset labeled with anomalous and normal classes based on the annotation methodology described in\citep{Brockmann2024}.
Anomalous test samples were generated by deliberately inducing errors during pick-and-place operations that commonly occur in operational actions, system components, and environmental conditions.
In contrast, normal samples were collected during undisturbed pick-and-place operations.
Since our method follows an unsupervised learning approach, all training data exclusively consist of samples from normal operations.

\begin{table}[t]
\footnotesize
\centering
\caption{List of anomaly types categorized by cause \citep{Brockmann2024}.}
\begin{tabular}{llr}
\toprule
\textbf{Anomaly type}       & \textbf{Cause}           & \multicolumn{1}{l}{\textbf{No.}} \\ \midrule
Additional friction         & \textit{Robot Axis Wear} & 144                              \\
Miscommutation              & \textit{Robot Axis Wear} & 89                               \\
Misgrip of can              & \textit{Gripping Errors} & 11                               \\
Losing the can              & \textit{Gripping Errors} & 74                               \\
Additional axis weight      & \textit{Process Errors}  & 156                              \\
Collision with foam         & \textit{Process Errors}  & 72                               \\
Collision with cables       & \textit{Process Errors}  & 48                               \\
Collision with cardboard    & \textit{Process Errors}  & 22                               \\
Varying can weight          & \textit{Process Errors}  & 80                               \\
Cable routed at robot       & \textit{Process Errors}  & 10                               \\
Invalid gripping position   & \textit{Process Errors}  & 12                               \\
Unstable Platform           & \textit{Process Errors}  & 37                               \\ \bottomrule
\end{tabular}
\label{tab:anomalies}
\end{table}

We modified the dataset to include only torque signals in order to meet our experimental requirements.
Since our model is designed to accurately operate with minimal sensor signals, we retained only a few signals that can be adopted in low-cost commercial robots.
In previous studies, torque signals have been commonly used to detect anomalies due to their strong indication of faults in the motor and the kinematic chain such as collisions or deterioration of robot joints \citep{Izagirre2020, Kermenov2023, Lou2019}.
Similarly, most anomalies in voraus-AD, such as collisions with objects or misgripping the can, induce fluctuations in torque values.
Therefore, we evaluated models using only 12 torque signals (2 torque sensors for each of the 6 joints) to accurately evaluate the anomaly detection performance under the constraint of using a limited number of sensor signals.
Our experiment shows that using only torque signals for comparison is valid, as torque values produced the best results across various models compared to other signals (Section \ref{sec:experiments:ablation:alternative}).
Figure \ref{fig:anomaly_normal_torques} illustrates torque signals for both anomalous and normal operations within a single sliding window from the 50 Hz dataset.

Additionally, we transformed the dataset to obtain multiple sampling frequencies: 100 Hz, 50 Hz, 25 Hz, and 10 Hz.
This is because robots can operate at various sampling rates, making it important to compare models across different frequencies to determine their applicability to a wide range of robots.
All modifications to the dataset were conducted using official implementation codes of the voraus-AD dataset.
Specifically, we used the 100 Hz dataset as the default and obtained the 50 Hz, 25 Hz, and 10 Hz datasets by uniformly downsampling it, selecting every $n$-th frame with $n = 100 / f_{\mathrm{target}}$, where $f_{\mathrm{target}} \in \{50, 25, 10\}$.

\begin{figure*}[t]
    \centering
    \includegraphics[width=0.85\linewidth]{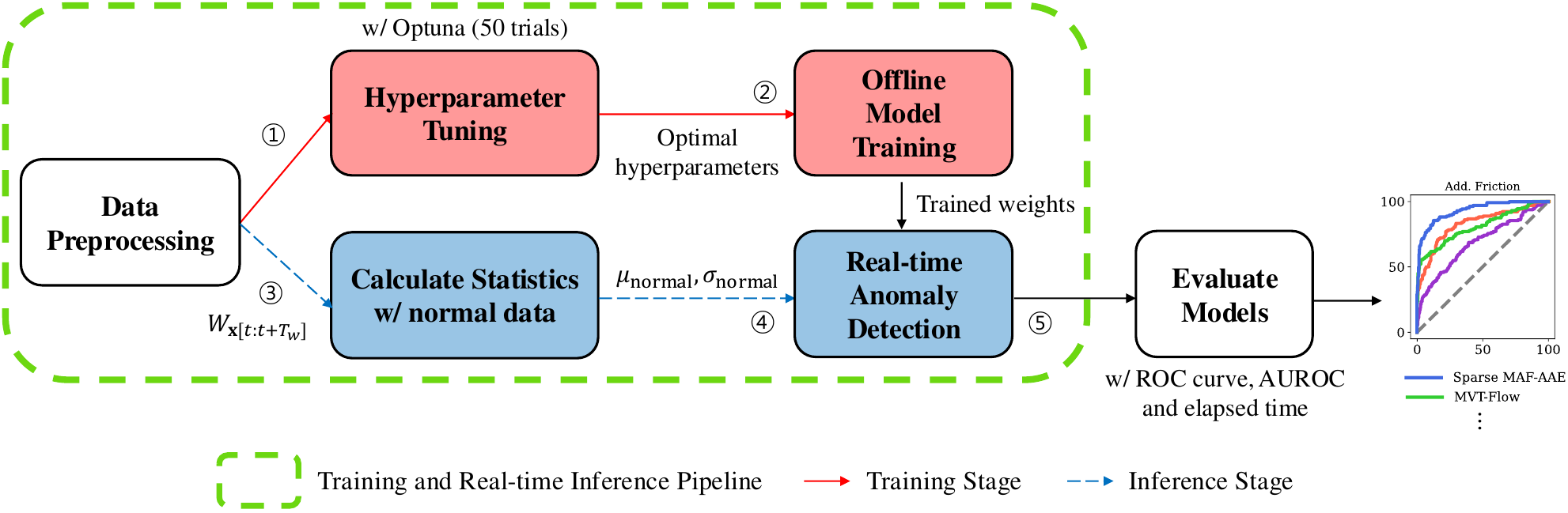}
    \caption{
    Training and inference pipeline in our experiments. All models followed the procedure outlined in the green dashed box and were compared in terms of performance and inference time.}
    \label{fig:experiment_procedure}
\end{figure*}

\subsection{Evaluation Processes and Metrics}\label{sec:experiments:eval_method}
We compared four anomaly detection models: Anomaly transformer \citep{Xu2022}, LSTM-VAE \citep{Park2018}, MVT-Flow -- the state-of-the-art on the voraus-AD dataset \citep{Brockmann2024}--, and Sparse MAF-AAE.
Since these models are based on unsupervised learning, training data can be readily collected by operating the robots under normal conditions.
Therefore, we compared the models using various dataset sizes—12.5\%, 25\%, 50\%, and 100\% of the training dataset—for both 100 Hz and 50 Hz sampling frequencies to observe how their performance changes with an increased amount of training data.

The overall experimental procedure was as follows: First, we optimized the hyperparameters for each model using the the Optuna framework \citep{Akiba2019}.
Next, we trained the models with these hyperparameters and evaluated each model 20 times, and then averaged the metrics.
Each model is assessed in two aspects: model performance, using the \textit{receiver operating characteristic} (ROC) curve and \textit{the area under its curve} (AUROC) expressed as a percentage, and elapsed time during inference to determine applicability for real-time tasks.
An overview of this procedure is illustrated in Figure \ref{fig:experiment_procedure}.

For real-time anomaly detection, we applied a sliding window technique to segment time-series data.
This method involved slicing the time-series data into segments of window size $T_W$ and moving to the next segment with a stride of $T_s$.
For example, if $T_W$ is $3$ seconds and $T_s$ is $0.5$ seconds, the initial window will cover the interval $[0.0, 3.0]$, followed by the next window $[0.5, 3.5]$.

Using the method described in Section \ref{sec:method:offline_training} and Section \ref{sec:method:real_time_detection}, the models were trained to accurately reconstruct signals within each window $W_{\mathbf{x}[t:t+T_W]}$.
They were then evaluated on their performance in discriminating whether the window contained an anomaly by reconstructing the input test window-sized data $\mathbf{w}$.
For each data sample, the window with the highest anomaly score was selected as the representative of the data.
Then, the models were evaluated by generating ROC curves and calculating AUROC percentages based on this set of highest scores.
The AUROC was measured for each anomaly type, and the overall performance was computed by averaging the values.
This approach mitigates the influence of varying sample sizes per anomaly type on the metrics, as discussed in \citep{Brockmann2024}.
To assess the elapsed time during inference, we calculated the average time between the interquartile range, which is between the first quartile (Q1) and the third quartile (Q3), to eliminate outliers.

\subsection{Implementation Details}\label{sec:experiments:implem_details}
All experiments were conducted using PyTorch \citep{Paszke2019} (v2.1.1), Python 3.11 on Ubuntu 22.04.
Model evaluations were performed on a desktop equipped with an NVIDIA GeForce RTX 4070 Ti GPU, Intel i9-14900KF CPU and 32GB RAM.
For MVT-Flow and Anomaly transformer, we used the code from their official GitHub repositories.
For LSTM-VAE, our own implementation was used.
Hyperparameters were divided into two groups: general hyperparameters for all models and specific hyperparameters for each model.

General hyperparameters included a batch size of 8, the AdamW \citep{Loshchilov2019} optimizer, and the MultipleStepLR learning rate scheduler.
The models were trained for 15 epochs with a learning rate decay at each epoch 2 and 12.
We set $T_W = 150$ and $T_s = 50$ for both 100 Hz and 50 Hz.
Thus, models with 100 Hz data performed anomaly detection using 1.5-second windows and a 0.5-second stride, while the models with 50 Hz data used 3-second windows with a 1-second stride.
For 25 Hz and 10 Hz, 3-second windows with a 1-second stride were used: $T_W = 75, T_s = 25$ for 25 Hz, and $T_W = 30, T_s = 10$ for 10 Hz data.

Specific hyperparameters, such as learning rate $\eta$ and decay $\gamma$, were optimized using the Optuna framework over 50 trials.
For Sparse MAF-AAE, we also optimized $\beta$ and $\lambda$, as discussed in Section \ref{sec:method:offline_training}.
The hidden and latent space sizes were doubled for Sparse MAF-AAE and halved for LSTM-VAE.
Anomaly scores for Sparse MAF-AAE and LSTM-VAE were calculated using the method outlined in Section \ref{sec:method:real_time_detection}, while MVT-Flow and the Anomaly transformer computed scores based on the $L_1$ norm scores.

\begin{table}[t]
\centering
\caption{Overall AUROC percentages for all models at update frequencies of 100 Hz, 50 Hz, 25 Hz, and 10 Hz.}
\scriptsize
\begin{tabular}{@{}lcccc@{}}
\toprule
\textbf{Model} & \begin{tabular}[c]{@{}c@{}}Anomaly\\ transformer\end{tabular} & LSTM-VAE      & MVT-Flow      & \begin{tabular}[c]{@{}c@{}}Sparse\\ MAF-AAE\end{tabular} \\ \midrule
100 Hz         & 57.23 ± 8.48                                                  & 73.66 ± 12.05 & 84.27 ± 11.55 & \textbf{89.23 ± 7.45}                                    \\
50 Hz          & 61.04 ± 5.46                                                  & 77.19 ± 10.80 & 77.39 ± 14.54 & \textbf{87.14 ± 8.69}                                    \\
25 Hz          & 62.18 ± 6.48                                                  & 78.21 ± 8.90  & 77.11 ± 14.78 & \textbf{81.97 ± 10.65}                                   \\
10 Hz          & 62.67 ± 8.46                                                  & 62.24 ± 13.33 & 72.39 ± 15.73 & \textbf{77.71 ± 10.48}                                   \\ \bottomrule
\end{tabular}
\label{tab:auroc_frequency}
\end{table}

\begin{table*}[t]
\scriptsize
\centering
\caption{AUROC percentages of online anomaly detection models on 100 Hz and 50 Hz torque signals \\trained with 948 (100\%) samples.}
\begin{tabular}{@{}l|cccc|cccc@{}}
\toprule
\textbf{Frequency} & \multicolumn{4}{c|}{100 Hz} & \multicolumn{4}{c}{50 Hz} \\ \midrule
\textbf{Model} & \begin{tabular}[c]{@{}c@{}}Anomaly\\ transformer\end{tabular} & LSTM-VAE & MVT-Flow & \begin{tabular}[c]{@{}c@{}}Sparse\\ MAF-AAE\end{tabular} & \begin{tabular}[c]{@{}c@{}}Anomaly\\ transformer\end{tabular} & LSTM-VAE & MVT-Flow & \begin{tabular}[c]{@{}c@{}}Sparse\\ MAF-AAE\end{tabular} \\ \midrule
Add. Friction & 69.63 & 81.60 ± 1.52 & 87.20 & 96.29 ± 0.03 & 68.81 & 83.99 ± 1.03 & 80.99 & 93.24 ± 0.08 \\
Miscommutation & 58.35 & 67.53 ± 1.75 & 90.96 & 75.87 ± 0.11 & 58.87 & 69.38 ± 1.99 & 83.90 & 74.86 ± 0.24 \\
Misgrip can & 62.53 & 97.99 ± 2.28 & 99.74 & 99.82 ± 0.02 & 68.37 & 97.92 ± 0.70 & 98.00 & 99.83 ± 0.02 \\
Losing can & 56.83 & 88.48 ± 1.35 & 88.20 & 89.03 ± 0.11 & 52.78 & 91.75 ± 1.04 & 82.23 & 89.04 ± 0.19 \\
Add. axis weight & 59.28 & 77.10 ± 1.34 & 88.50 & 84.55 ± 0.06 & 65.81 & 73.37 ± 1.09 & 79.82 & 82.39 ± 0.15 \\
Coll. foam & 64.45 & 61.79 ± 2.14 & 68.22 & 88.08 ± 0.07 & 64.44 & 69.22 ± 1.71 & 68.16 & 77.18 ± 0.16 \\
Coll. cables & 52.17 & 59.55 ± 2.80 & 66.00 & 82.86 ± 0.19 & 51.68 & 61.74 ± 2.29 & 52.51 & 72.89 ± 0.26 \\
Coll. cardboard & 55.44 & 56.18 ± 4.64 & 68.87 & 91.17 ± 0.14 & 58.19 & 70.44 ± 2.78 & 57.26 & 91.44 ± 0.33 \\
Var. can weight & 53.71 & 69.28 ± 1.95 & 82.66 & 80.29 ± 0.10 & 59.32 & 83.16 ± 1.32 & 68.68 & 84.75 ± 0.21 \\
Cable at robot & 34.99 & 76.52 ± 5.25 & 100.00 & 98.89 ± 0.04 & 62.94 & 83.96 ± 3.28 & 100.00 & 93.79 ± 0.22 \\
Invalid grip. pos. & 58.63 & 74.71 ± 6.05 & 90.97 & 89.29 ± 0.14 & 62.05 & 72.16 ± 3.27 & 86.22 & 95.64 ± 0.18 \\
Unstable platform & 60.80 & 73.23 ± 2.51 & 79.97 & 94.65 ± 0.08 & 59.25 & 69.16 ± 1.70 & 70.85 & 90.66 ± 0.19 \\ \midrule
\textbf{Mean ± SD} & \textbf{57.23 ± 8.48} & \textbf{73.66 ± 12.05} & \textbf{84.27 ± 11.55} & \textbf{89.23 ± 7.45} & \textbf{61.04 ± 5.46} & \textbf{77.19 ± 10.80} & \textbf{77.39 ± 14.54} & \textbf{87.14 ± 8.69} \\ \bottomrule
\end{tabular}
\begin{flushleft}
The mean represents the average score across all anomaly types, and SD stands for the standard deviation among all anomaly types.\\ 
The variance for each type of result produced by the Anomaly transformer and MVT-Flow models is 0.0, as these models are deterministic in nature.\\
Consequently, only the mean values are reported for each type in these models.
\end{flushleft}
\label{tab:auroc_1.0}
\end{table*}

\subsection{Results}\label{sec:experiments:results}
In this subsection, the experimental results for anomaly detection performance using torque signals from the voraus-AD dataset sampled at various frequencies are presented.
The models were also evaluated on various training dataset sizes to determine if their performance can be enhanced by collecting more normal operational data.
Furthermore, the elapsed time for inference of each model on the 50 Hz sampling frequency was compared.
It is recommended to view all figures in this section in color print.

As listed in Table~\ref{tab:auroc_frequency}, our model achieved \textit{state-of-the-art} results at all frequencies using torque signals in the voraus-AD dataset.
Sparse MAF-AAE demonstrated significant advantages, particularly in the 100 Hz and 50 Hz results.
It outperformed other models by a substantial margin, achieving 4.96\% higher performance than MVT-Flow, the SOTA model for the original voraus-AD dataset, and 15.57\% higher than LSTM-VAE.
For the 50 Hz results, Sparse MAF-AAE also surpassed others, with a notable margin of 9.75\% over MVT-Flow.
These results highlight the superior efficacy of our model in detecting anomalies across different frequencies.

\begin{figure*}[tbp]
    \centering
    \includegraphics[width=\linewidth]{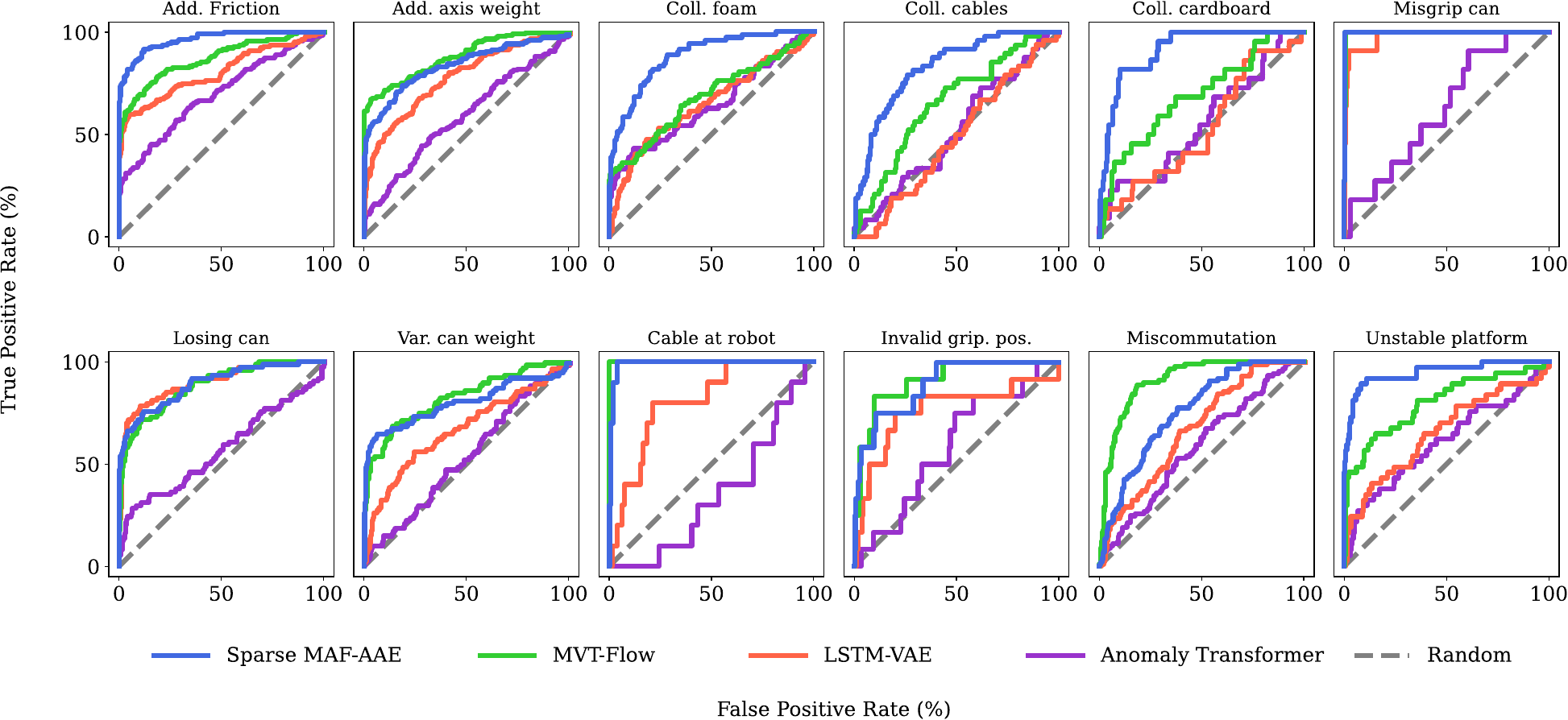}
    \caption{ROC curve for each anomaly type at 100 Hz data.}
    \label{fig:roc_100hz_1.0}
\end{figure*}

\begin{figure*}[tbp]
    \centering
    \includegraphics[width=\linewidth]{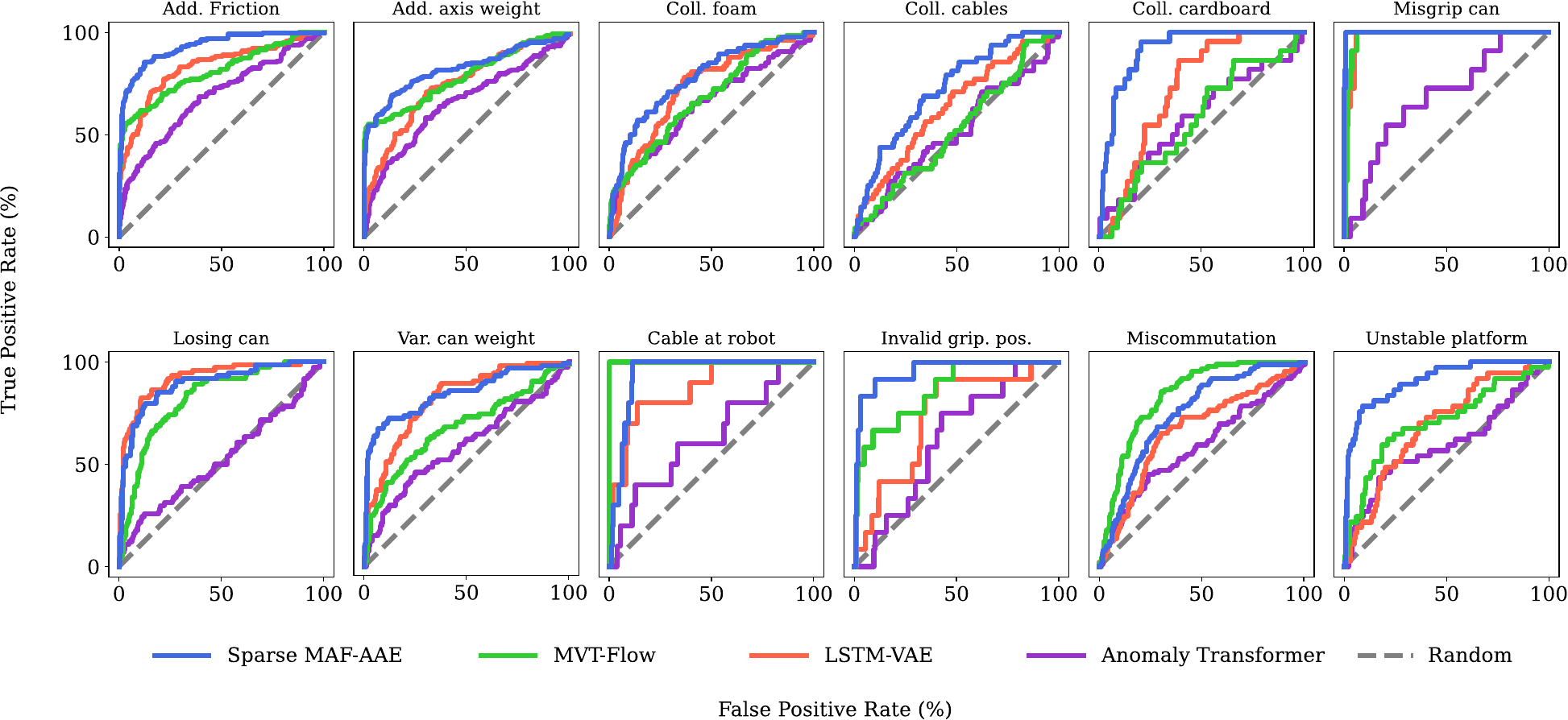}
    \caption{ROC curve for each anomaly type at 50 Hz data.}
    \label{fig:roc_50hz_1.0}
\end{figure*}

For a detailed comparison, we calculated the AUROC percentages of each anomaly type and the average performance of each model on 100 Hz and 50 Hz torque signals, respectively, trained with 948 (100\%) samples.
Figure \ref{fig:roc_100hz_1.0} and Figure \ref{fig:roc_50hz_1.0} showcase the ROC curves for each initial run on 100 Hz and 50 Hz, respectively.
The numerical results are presented in Table \ref{tab:auroc_1.0}.
Sparse MAF-AAE outperformed others in 7 out of 12 categories at 100 Hz and 9 out of 12 at 50 Hz.
Furthermore, our model showed the smallest variance across all anomalies when compared to LSTM-VAE and MVT-Flow.
This consistency indicates that our model not only excels in detecting a wide range of anomalies but also maintains superior and uniform performance across all categories, underscoring its robustness and reliability.

Specifically, Sparse MAF-AAE excels in detecting collisions with lightweight objects.
Collision detection includes collision with foam, cables, and cardboard.
Table \ref{tab:auroc_collision} presents AUROC percentages for collision detection of each model.
Sparse MAF-AAE surpasses previous works by a large margin, achieving 19.67\% higher performance on 100 Hz signals and 13.37\% higher on 50 Hz signals.
Considering that torques are primarily affected during collisions, our model demonstrates superior effectiveness in detecting anomalous fluctuations in torque signals compared to existing models.
Moreover, since collisions with lightweight objects are critical for robot operations but do not produce significant torque gradients, the result indicates that our model excels in detecting subtle yet crucial errors compared to others.
Our model did not perform well in detecting miscommutation, likely because miscommutation does not have a direct relationship with torque values \citep{Brockmann2024}.
This indicates that the model correctly identifies anomalies related to torque values and does not produce false positives for anomalies outside its scope.

\begin{figure}[t]
    \centering
    \includegraphics[width=\linewidth]{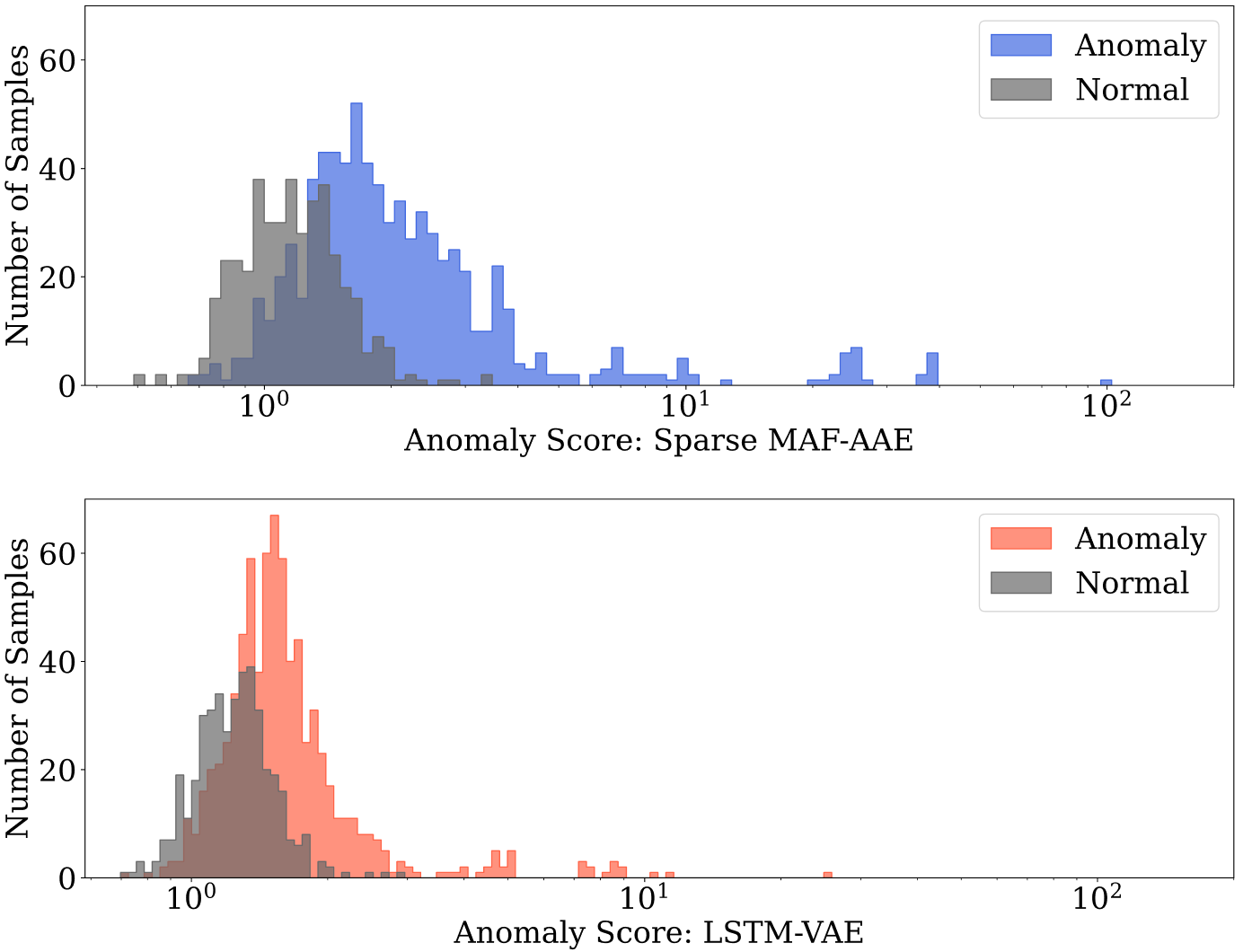}
    \caption{Log-scale anomaly scores of Sparse MAF-AAE and LSTM-VAE.
    We selected LSTM-VAE as a comparison model because it calculates anomaly scores in the same way as Sparse MAF-AAE.}
    \label{fig:anomaly_scores}
\end{figure}

Figure \ref{fig:anomaly_scores} visualizes the distribution of log-scale anomaly scores from Sparse MAF-AAE and LSTM-VAE on the 50 Hz dataset.
The histogram demonstrates that Sparse MAF-AAE has fewer regions of overlap between the distributions of anomalies and normals compared to LSTM-VAE, indicating its capacity to distinctly separate these classes.
Also, the long-tailed distribution of anomaly scores in Sparse MAF-AAE highlights its proficiency in assigning appropriately higher scores to anomalies, while normal scores seem to follow a normal distribution, reflecting the model's robustness in distinguishing anomalous data.


\begin{table}[t]
\centering
\caption{AUROC percentages for collision detection using torque signals.}
{\scriptsize
\begin{tabular}{@{}lcccc@{}}
\toprule
\textbf{Model} & \begin{tabular}[c]{@{}c@{}}Anomaly\\ transformer\end{tabular} & LSTM-VAE & MVT-Flow & \begin{tabular}[c]{@{}c@{}}Sparse\\ MAF-AAE\end{tabular} \\ \midrule
100 Hz & 57.35 ± 6.36 & 59.17 ± 2.82 & 67.70 ± 1.50 & \textbf{87.37 ± 4.20} \\
50 Hz & 58.10 ± 6.38 & 67.13 ± 4.71 & 59.31 ± 8.02 & \textbf{80.50 ± 9.71} \\
25 Hz & 60.51 ± 3.56 & 70.29 ± 4.78 & 58.90 ± 3.33 & \textbf{71.50 ± 7.76}\\ 
10 Hz & 60.81 ± 2.10 & 53.68 ± 3.60 & 56.11 ± 3.64 & \textbf{68.61 ± 5.29}\\ \bottomrule
\end{tabular}
}
\label{tab:auroc_collision}
\vspace{-1em}
\end{table}

\begin{figure}[ht]
    \centering
    \includegraphics[width=\linewidth]{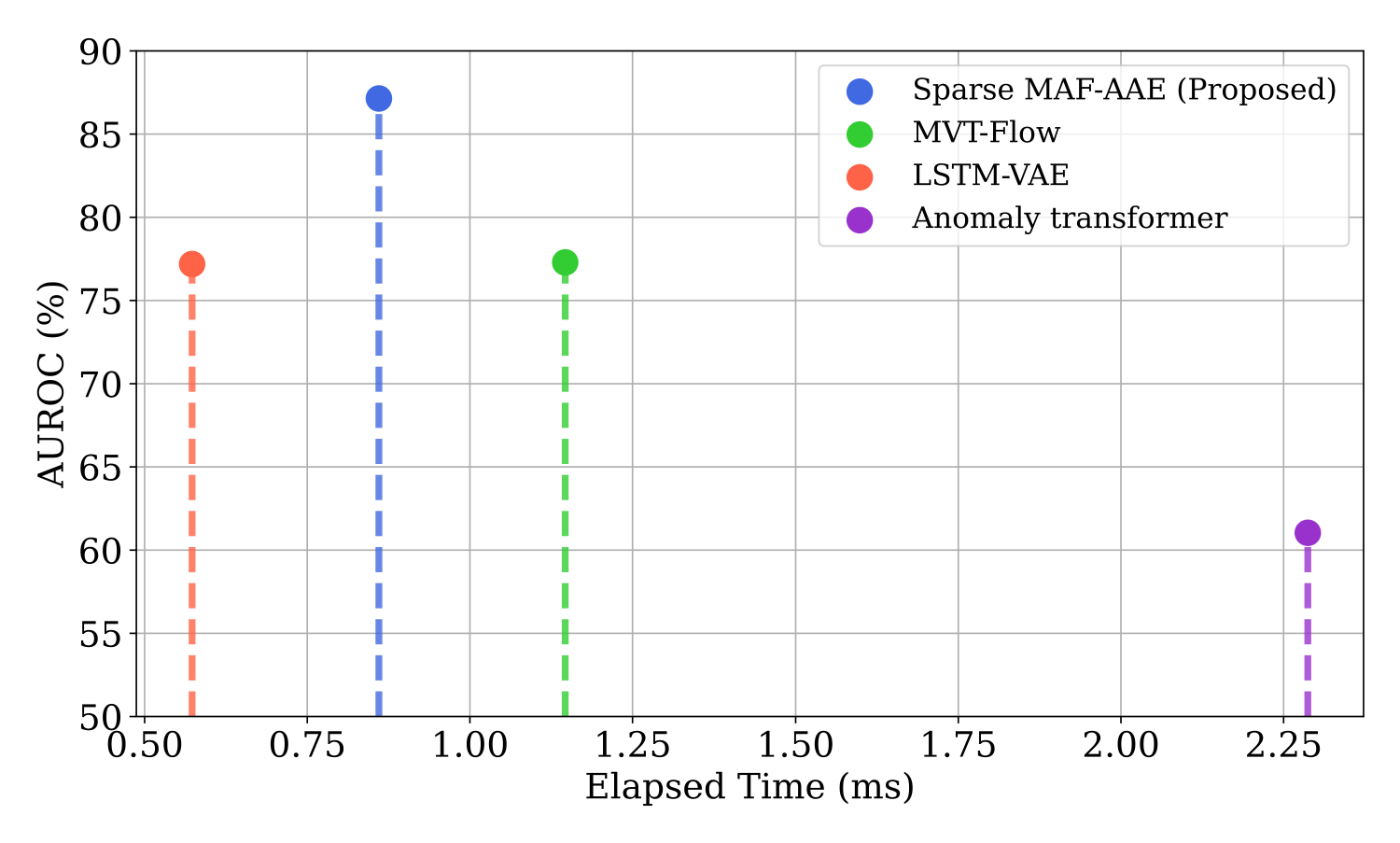}
    \caption{Average elapsed time for inference versus AUROC percentages for all   
             baselines on the 50 Hz dataset.
             Details of the evaluation environment can be found in Section \ref{sec:experiments:implem_details}.}
    \label{fig:elapsed_time}
\end{figure}

Anomaly transformer, which has achieved success in various scenarios \citep{Xu2022}, showed the worst performances among the models in this experiment.
These are likely due to the inherent requirement of transformers for large datasets to achieve effective training in many domains \citep{Shao2022, Xu2021}.
These results indicate that in the robotics domain, where collecting sufficient data is limited by time and resources \citep{Soori2023}, large models with many parameters may be impractical.

Figure \ref{fig:elapsed_time} shows both the average elapsed time required to reconstruct the data and calculate the anomaly score, as well as the overall model performance on the 50 Hz dataset.
Although all baselines were fast enough for real-time anomaly detection, LSTM-VAE (0.57 ms) and Sparse MAF-AAE (0.86 ms) performed in less than 1 ms, while MVT-Flow (1.15 ms) and Anomaly transformer (2.29 ms) did not meet the 1ms threshold.
Despite its low computational load, Sparse MAF-AAE demonstrated outstanding detection performance compared to other baselines, with a significant margin of about 10\%. 
This makes Sparse MAF-AAE exceptionally suitable for real-time robotic applications, where rapid and accurate anomaly detection is essential to prevent failures and ensure continuous operations.

\begin{figure*}[tbp]
  \centering
  \includegraphics[width=\linewidth]{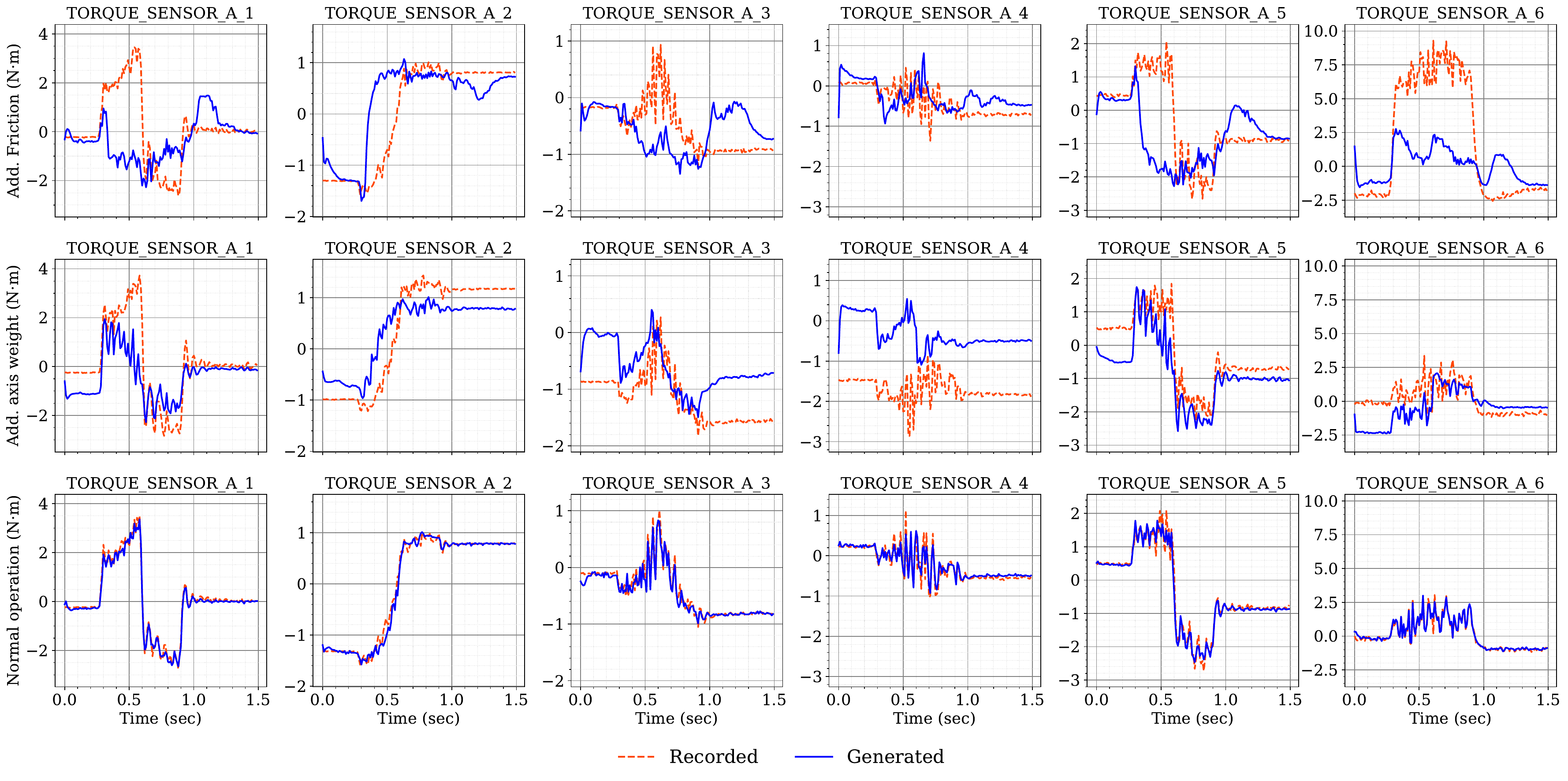}
  \caption{
  Comparison of recorded (input) and generated (output) 50 Hz torque signals by Sparse MAF-AAE under anomalous conditions (additional friction, additional axis weight) and normal operation, within the same 1.5-second sliding window. Each joint is equipped with two torque sensors (Sensor A and Sensor B); only Sensor A is shown in this visualization.}
  \label{fig:torque_plots}
\end{figure*}
\begin{figure*}[tbp]
  \centering
  \includegraphics[width=\linewidth]{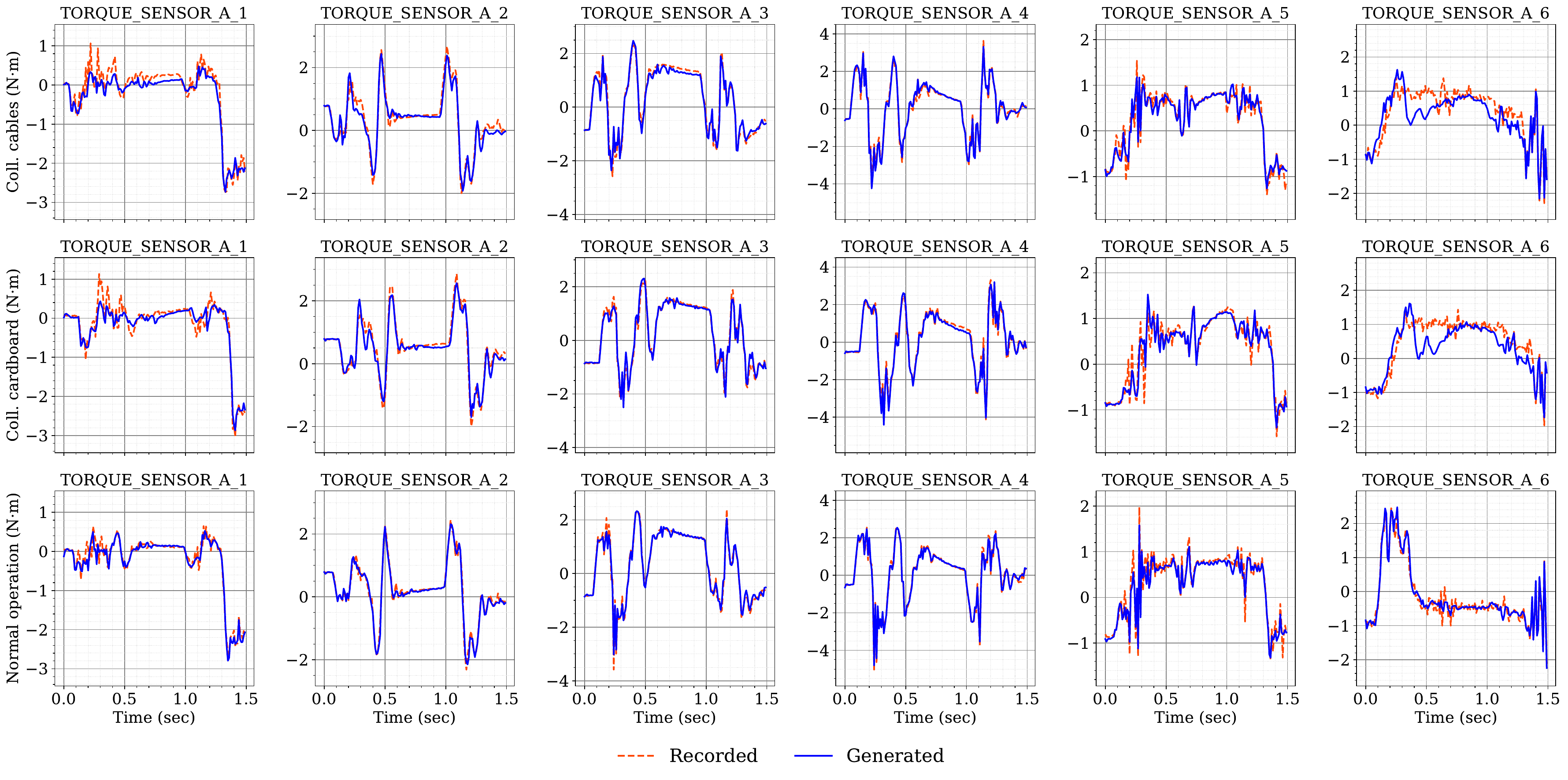}
  \caption{
  Comparison of recorded (input) and generated (output) 50 Hz torque signals by Sparse MAF-AAE under lightweight collision scenarios (collision with cables, collision with cardboard) and normal operation, within the same 1.5-second sliding window. Each joint is equipped with two torque sensors (Sensor A and Sensor B); only Sensor A is shown in this visualization.}
  \label{fig:torque_plots_collision}
\end{figure*}

For qualitative evaluation, we plot the recorded and generated torque signals from the 50 Hz dataset to verify that our model accurately reconstructs normal signals, while failing to do so for anomalous inputs.
As shown in Figures \ref{fig:torque_plots} and \ref{fig:torque_plots_collision}, our model precisely reproduces normal signals but exhibits discrepancies under various anomalies, resulting in high anomaly scores.
Notably, Figure \ref{fig:torque_plots_collision} demonstrates the model’s ability to produce distinguishable deviations from recorded signals even under subtle anomalous conditions, such as collisions with lightweight objects that affect only a subset of sensors (torque sensors 1 and 6).
This visualization shows that our model amplifies signal deviations under anomalies while accurately reconstructing normal signals, enabling effective anomaly detection.

We also investigated whether increasing the amount of training data can improve model performance.
Table \ref{tab:auroc_traingain} presents AUROC percentages for each training set size on the 50 Hz dataset.
Sparse MAF-AAE demonstrated the best performance across most training set sizes, indicating that the model is robustly applicable to scenarios with varying amounts of training data.
Furthermore, as the amount of training data increases, Sparse MAF-AAE showed potential for further performance improvements.
Collecting training data for baselines (i.e., unsupervised learning models) is relatively straightforward than for supervised learning approaches, as it can be easily gathered by simply operating robots under normal conditions.
This scalability makes our model an ideal choice for long-term deployment in robotic systems.

\subsection{Ablation Studies}\label{sec:experiments:ablation}
To thoroughly understand the characteristics of our model, we conducted additional experiments from three perspectives: eliminating the layers and components we added, using selected signals for anomaly detection, and employing all machine signals from voraus-AD as done by \citep{Brockmann2024} in their experiments.

\subsubsection{Effects of Two Main Components in Sparse MAF-AAE}\label{sec:experiments:ablation:main_comp}
\begin{table}[t]
\centering
\scriptsize
\caption{AUROC percentages for each training set size on the 50 Hz dataset.}
\begin{tabular}{@{}lcccc@{}}
\toprule
\textbf{Model} & \multicolumn{1}{c}{\begin{tabular}[c]{@{}c@{}}Anomaly\\ transformer\end{tabular}} & \multicolumn{1}{c}{LSTM-VAE} & \multicolumn{1}{c}{MVT-Flow} & \multicolumn{1}{c}{\begin{tabular}[c]{@{}c@{}}Sparse\\ MAF-AAE\end{tabular}} \\ \midrule
100\%          & 61.04 ± 5.46                                                                      & 77.19 ± 10.80                & 77.39 ± 14.54                & \textbf{87.14 ± 8.69}                                                        \\
50\%           & 60.13 ± 8.92                                                                      & 79.58 ± 9.90                 & 77.27 ± 14.03                & \textbf{82.65 ± 8.39}                                                        \\
25\%           & 55.75 ± 5.26                                                                      & 63.79 ± 12.06                & \textbf{75.76 ± 9.03}        & 73.55 ± 10.36                                                                \\
12.5\%         & 58.12 ± 5.57                                                                      & 64.11 ± 11.46                & 75.25 ± 9.46                 & \textbf{76.24 ± 10.96}                                                       \\ \bottomrule
\end{tabular}
\label{tab:auroc_traingain}
\end{table}
\begin{figure}[t]
    \centering
    \includegraphics[width=\linewidth]{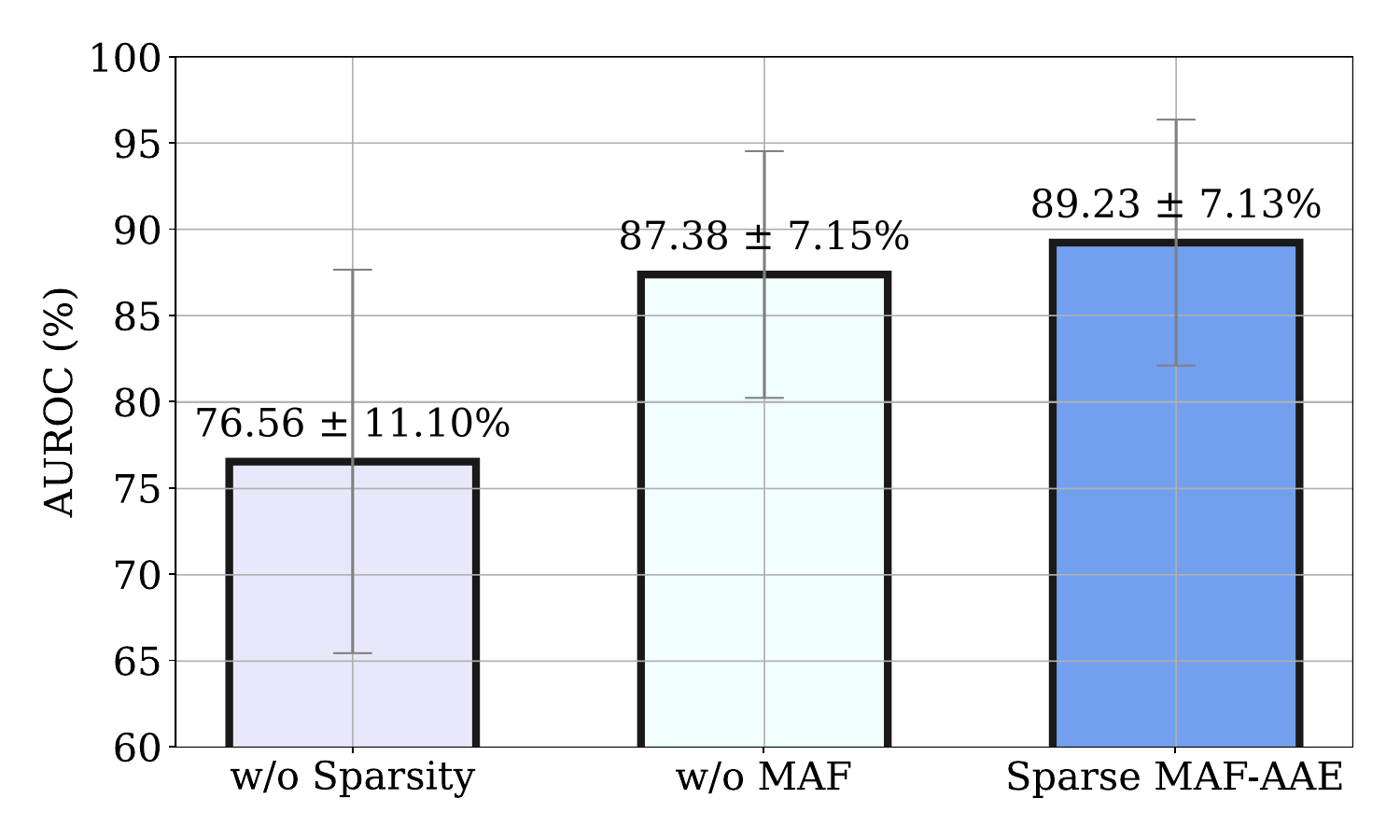}
    \caption{AUROC percentages of Sparse MAF-AAE and models without each improvement.
            The models were compared using torque signals from 100 Hz frequency dataset.}
    \label{fig:eliminate_layers}
\end{figure}

In this study, we analyzed the impact of two principal components proposed in Sparse MAF-AAE: the sparsity constraints and the MAF layers on AAE encoders.
We evaluated the impact of sparsity by configuring the model to physically compress the input space to construct the latent space, as is common practice in typical autoencoders.
To assess the effect of MAF layers, we constrained the posterior and prior distributions to follow a Gaussian distribution.
After removing specific layers, hyperparameter tuning for these models was conducted using the same methodology outlined in Section \ref{sec:experiments:eval_method} to ensure a fair comparison.

As illustrated in Figure \ref{fig:eliminate_layers}, each layer we added significantly enhanced the anomaly detection performance of Sparse MAF-AAE.
The results indicate that activating important cells using mathematical constraints is more effective than the model in limiting the size of the latent space when the input data have a small number of features.
This approach is more practical for industrial applications because it enables precise inferences even with a limited number of sensor signals.
Also, the results show that transforming the latent space distribution enhances the model's ability to accurately distinguish anomalies and ensures robust performance with minimal deviation across different anomaly types.
Therefore, the components we added and changed significantly improve the model's accuracy in detecting anomalies across all categories, making it exceptionally effective for diverse and challenging anomaly detection tasks.

\subsubsection{Using Selected Signals for Anomaly Detection}\label{sec:experiments:ablation:alternative}
In this study, whether each selected signal type can effectively represent anomalies was evaluated.
We selected 6 encoder values (joint positions), 6 torque-forming current values, and 12 torque values as the features of the input data.
MVT-Flow and Sparse MAF-AAE were employed to assess the validity of each dataset.

\begin{table}[t]
\footnotesize
\centering
\caption{Overall AUROC percentages using selected signals on the 100 Hz dataset.}
\begin{tabular}{@{}lcc@{}}
\toprule
\textbf{Model} & MVT-Flow               & Sparse MAF-AAE        \\ \midrule
Encoders       & 63.09 ± 13.99          & 61.26 ± 13.94         \\
Currents       & 74.30 ± 12.57          & 73.79 ± 14.08         \\
Torques        & \textbf{84.27 ± 11.55} & \textbf{89.23 ± 7.45} \\ \bottomrule
\end{tabular}
\label{tab:selected_all}
\end{table}

\begin{table}[t]
\footnotesize
\centering
\caption{AUROC percentages for miscommutation using selected signals on the 100 Hz dataset.}
\begin{tabular}{@{}lcc@{}}
\toprule
\textbf{Model} & MVT-Flow              & Sparse MAF-AAE        \\ \midrule
Encoders       & 58.13          & 54.03 ± 0.65          \\
Currents       & \textbf{92.14} & \textbf{98.65 ± 0.03} \\
Torques        & 90.96          & 75.87 ± 0.11          \\ \bottomrule
\end{tabular}
\label{tab:selected_miscommutation}
\end{table}

As shown in Table \ref{tab:selected_all}, the dataset composed of torque signals achieved the highest overall AUROC percentages, indicating that torque signals are the most appropriate in representing anomalies among the selected signals.
Since both encoder and current signals demonstrate lower overall performance compared to torque signals in both models, it can be concluded that they are not effective in representing various anomalies on their own.
These results highlight the importance of selecting input features that are highly relevant to the causes of anomalies.
Although anomalies can be detected using encoder or current values, our model’s performance is less robust in these cases due to their limited sensitivity to most types of anomalies.

While torque signals were shown to be the most effective overall, current signals showed the best performance in detecting miscommutation errors compared to other signals (Table \ref{tab:selected_miscommutation}).
Since current signals are primarily affected by miscommutation \citep{Brockmann2024}, this suggests that our experimental setup is well-constructed, as models consistently detected the anomaly with a clear causal relationship.
Furthermore, because the models demonstrated significantly better performance using only torque signals compared to other signals in this environment, it is justified to use torque signals as a representative measure for comparing model performance.

\subsubsection{Utilizing All Machine Signals}\label{sec:experiments:ablation:all}
In this experiment, we used all machine signals (i.e., all 130 signals without metadata) from voraus-AD to comprehensively evaluate model performance and validate our experiments by comparing AUROC percentages with the metrics reported from the study that proposed MVT-Flow \citep{Brockmann2024}.
In our experiments, using all machine signals with MVT-Flow on the 100 Hz dataset resulted in an AUROC score of 91.81\%.
This closely aligns with the previously reported performance of 93.60\% for MVT-Flow.
These results confirm the validity and reliability of our experimental setup and comparisons, as they successfully reproduced the original study's score.

Additionally, we compared the performance of the model when using only torque signals versus using all machine signals to detect anomalies.
Sparse MAF-AAE's performance declined when all machine signals were included (77.25\%), compared to using only torque signals (89.23\%).
It is expected that the model may not perform well when a large number of diverse features are used, since our model is designed to expand the size of the latent space to effectively detect anomalies with a limited number of signals.
However, performing well with a small number of signals is more advantageous from an engineering perspective, as it is more applicable in various robotic and industrial situations.
Therefore, our model demonstrates greater practicality by achieving a comparable AUROC percentage to MVT-Flow (91.81\%) while requiring significantly fewer signals.

\section{Conclusion}
In this study, we propose Sparse MAF-AAE, a model designed for real-time multivariate anomaly detection in commercial robot operations.
Its lightweight architecture, flexible latent space, and sparsity constraints enable effective performance with minimal sensor signals.
Sparse MAF-AAE also offers improved robustness against out-of-distribution problems compared to existing NF-based models, enhancing reliability in safety-critical applications.
Experiments with a modified voraus-AD dataset \citep{Brockmann2024}, reflecting scenarios for commercial robots, showed that our model significantly outperformed baseline models and demonstrated computational efficiency suitable for real-time anomaly detection.

This research highlights the potential of Sparse MAF-AAE for application in low-cost commercial robots, which may have limited sensors but require high safety standards.
The model ensures operational safety and reliability through high performance and real-time detection capabilities, enhancing task efficiency \citep{Kanazawa2019}.
Its robustness and efficiency extend its applicability beyond robotics to industries where safety, reliability, and productivity are crucial, such as manufacturing systems \citep{Kharitonov2022}.
Given these characteristics, Sparse MAF-AAE is expected to significantly contribute to safer and more efficient operational systems in both robotic and engineering domains.



 \bibliographystyle{elsarticle-harv} 
 \bibliography{cas-refs}





\end{document}